\documentclass[review]{elsarticle}

\usepackage{amssymb,amsmath,amsthm}
\usepackage{makecell}
\usepackage{subfigure}
\usepackage{eso-pic}
\usepackage{accents}
\usepackage{pict2e,xfp}
\usepackage{upgreek}

\makeatletter
\newcommand{\dvee}{\mathbin{\mathpalette\d@vee\relax}}
\newcommand{\d@vee}[2]{%
\begingroup
\sbox\z@{$\m@th#1\vee$}%
\setlength{\unitlength}{\ht\z@}%
\kern0.1\wd\z@
\begin{picture}(\fpeval{0.4\wd0/\unitlength},1)
\roundcap\d@vee@thickness{#1}
\Line(0,1.03)(\fpeval{0.4\wd0/\unitlength},0)
\end{picture}%
\kern-0.3\wd\z@\box\z@
\endgroup
}
\newcommand{\d@vee@thickness}[1]{
\linethickness{%
1\fontdimen8
\ifx#1\displaystyle\textfont\else
\ifx#1\textstyle\textfont\else
\ifx#1\scriptstyle\scriptfont\else
\scriptscriptfont\fi\fi\fi 3
}%
}
\makeatother

\begin{document}

\begin{frontmatter}

\title{When Kinematics Dominates Mechanics: Locally Volume-Preserving Primitives for Model Reduction in Finite Elasticity}


\author[mymainaddress]{Xu Yi}
\ead{x.yi@u.nus.edu}

\author[mymainaddress]{Gregory Scott Chirikjian\corref{mycorrespondingauthor}}
\cortext[mycorrespondingauthor]{Corresponding author}
\ead{mpegre@nus.edu.sg}

\address[mymainaddress]{Department of Mechanical Engineering, National University of Singapore, Singapore}

\begin{abstract}
A new, and extremely fast, computational modeling paradigm is introduced here for specific finite elasticity problems that arise in the context of soft robotics. Whereas continuum mechanics is a very classical area of study, and significant effort has been devoted to the development of intricate constitutive models for finite elasticity, we show that in the kinds of large-strain mechanics problems arising in soft robotics, many of the parameters in constitutive models are irrelevant. For the most part, the isochoric (locally volume-preserving) constraint dominates behavior, and this can be built into closed-form kinematic deformation fields before even considering other aspects of constitutive modeling. We therefore focus on developing and applying primitive deformations that each observe this constraint. It is shown that by composing a wide enough variety of such deformations that the most common behaviors observed in soft robots can be replicated. Case studies include an inflatable rubber chamber, a slender rubber rod, and a rubber block subjected to different boundary conditions. We show that this method is at least 50 times faster than the ABAQUS implementation of the finite element method (FEM). Physical experiments and measurements show that both our method and ABAQUS have approximately $10\%$ error relative to experimentally measured displacements, as well as to each other. Our method provides a real-time alternative to FEM, and captures essential degrees of freedom for use in feedback control systems.

\end{abstract}


\end{frontmatter}


\section{Introduction}

Continuum mechanics and finite elasticity are classical fields wherein many research topics are considered
solved, and acceptable answers to engineering problems can be provided by using a finite element method (FEM)
code. However, the area of soft robotics is an exception to this because to perform real-time feedback control
it is desirable for the computational model of the physical system to update tens or hundreds of times per second. This is obviously beyond the range of current FEM capabilities, and likely will be for some time to come.
We therefore propose an alternative modeling paradigm, akin to Galerkin methods in linear elasticity. But instead
of addition of weighted modes defined over the continuum, functional composition of large-strain deformation fields are considered. The reason for doing this is that if each primitive deformation is isochoric, then so too will be the composition, but not the sum. More details about this will follow, but first some background is provided.

Conventional robots consist of rigid links which feature precise control and highly reproducible trajectories. 
They have a finite number of degrees of freedom, are governed by ordinary differential equations (ODEs) and algorithms for their control are well known. In contrast,
an elastic continuum has an infinite number of degrees of freedom governed by partial differential equations (PDEs), and methods for control are not as well developed. 

In recent years, the design and modeling of soft robots has emerged as one of the most significant evolutions in the field of robotics. Soft robots are fabricated from highly pliable materials (e.g. plastics, silicone, textiles, or cables) \cite{rus2015design}, which allow the robots to reshape continuously and softly with simple actuation, hence outstripping their rigid counterpart in human interaction, adaptability and energy efficiency.
However, the high degrees of freedom and nonlinear material property of soft robots complicate the modeling and control work significantly. The lack of simulation or analysis tools has also become one of the major challenges in this research field \cite{lipson2014challenges}.

The majority of the current models used in soft robotics are mechanics-based such as FEM, which plays a significant role in soft structure research \cite{laschi2012soft, polygerinos2013towards}. These studies prove that FEM can handle complex simulations, such as structures with multiple of materials or structures that interact with the surroundings. Significant research efforts have been put on expediating FEM modeling: Duriez et al. \cite{duriez2017soft,duriez2013control} presented a real-time FEM with an optimization approach for the control of soft robots. Moseley et al. \cite{moseley2016modeling} also developed an open-source tool, which interfaced with commercial FEM software, in order to automate the simulation process.
FEM specializes in representing soft robot configurations precisely since it follows closely to the principle laws, while providing researchers sufficient degrees of freedom (DOFs), but it also involves extensive computational processes due to its complexity. The application of FEM to real-time control systems is not practical because it is too slow.

When the system is not too complex, researchers also develop their own models to capture the motion of soft actuators. Polygerinos et al. \cite{polygerinos2015modeling} modeled the principle of operation of fiber-reinforced bending actuators with a quasi-static model; Jones et al. \cite{jones2009three} used Cosserat rod theory to capture the configuration of a continuum robot; Della Santina et al. \cite{della2020model} modeled a planer bending actuator based on the piecewise constant curvature hypothesis. These modeling methods have high accuracy, but also require complex computation.

Methods without constitutive laws were also developed. Computational algorithms analyze data obtained from physical experiments to generate the optimal deformation parameters. There are geometry-based approaches such as Fang's work \cite{fang2018geometry}, which takes geometric actuation as input and uses a numerical optimization algorithm to compute the deformation of soft robots. In recent years, methods involving machine learning have also become increasingly prevalent. Neural network approaches \cite{gillespie2018learning,thuruthel2019soft} formulate the discrete state-space representation of the system, which enabled position control within a small deformation. Deep learning \cite{truby2020distributed} and regression analysis \cite{elgeneidy2018bending} derive the relationship between bending angle and input pressure.
Computational methods skip the tedious calculation of energy minimization, but they rely on a large number of input data and repeated training processes.

\begin{figure*}[h]
\centering
\includegraphics[scale=0.4]{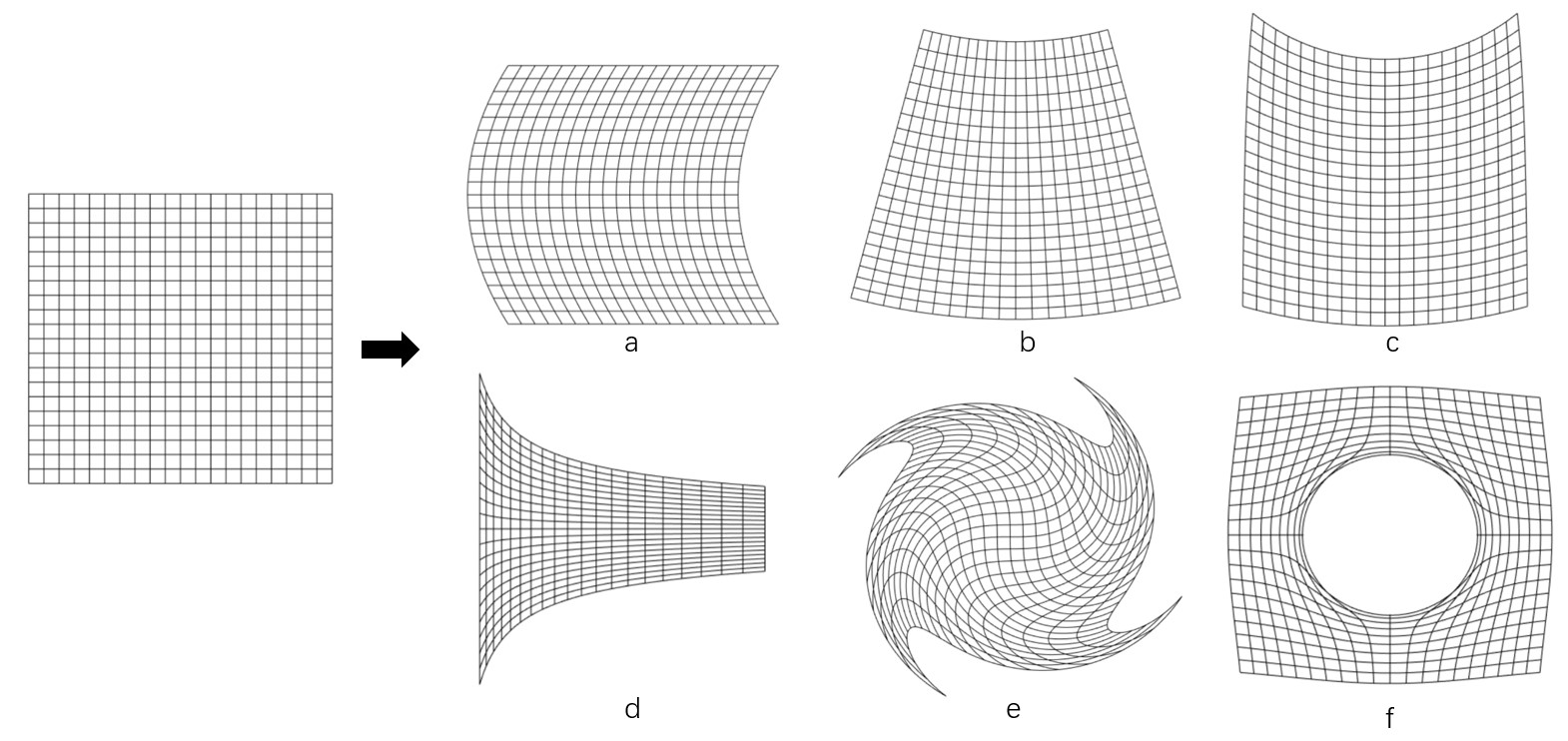}
\caption{Locally volume preserving deformation primitives: a. shear; b. variable offset bending; c. offset shear bending; d. elongation and contraction; e. votex; f. source \cite{chirikjian1995closed}.}
\label{lvpd}
\end{figure*}

This work aims to find a simple method to use kinematic deformation primitives applied to soft structures. In real life, the incompressibility assumption holds for rubber-like materials. Hence, every differential volume element of the elastic material could be treated as volume-preserving. The deformation can then be modeled by local volume-preserving deformations as in \cite{chirikjian1995closed}, where several closed-form deformation primitives were defined. Figure \ref{lvpd} illustrates several primitives. Basic scaler functions that can be expressed as a weighted sum of modes are incorporated in these primitives to provide DoFs for individual deformations.
These primitives are then combined with some optimization methods such as Newton's method, Jacobian method, etc. The optimization drives the parameters in scaler functions to satisfy the boundary conditions of the system. Unlike FEM, this method simulates the robot deformation at a global level. This cuts down the number of DOFs, which enables rapid simulation compared to the FEM method. It also saves the prolonged data collection and training process that machine learning methods require.

\section{Method}
\begin{figure}[ht]
\centering
\includegraphics[scale=0.3]{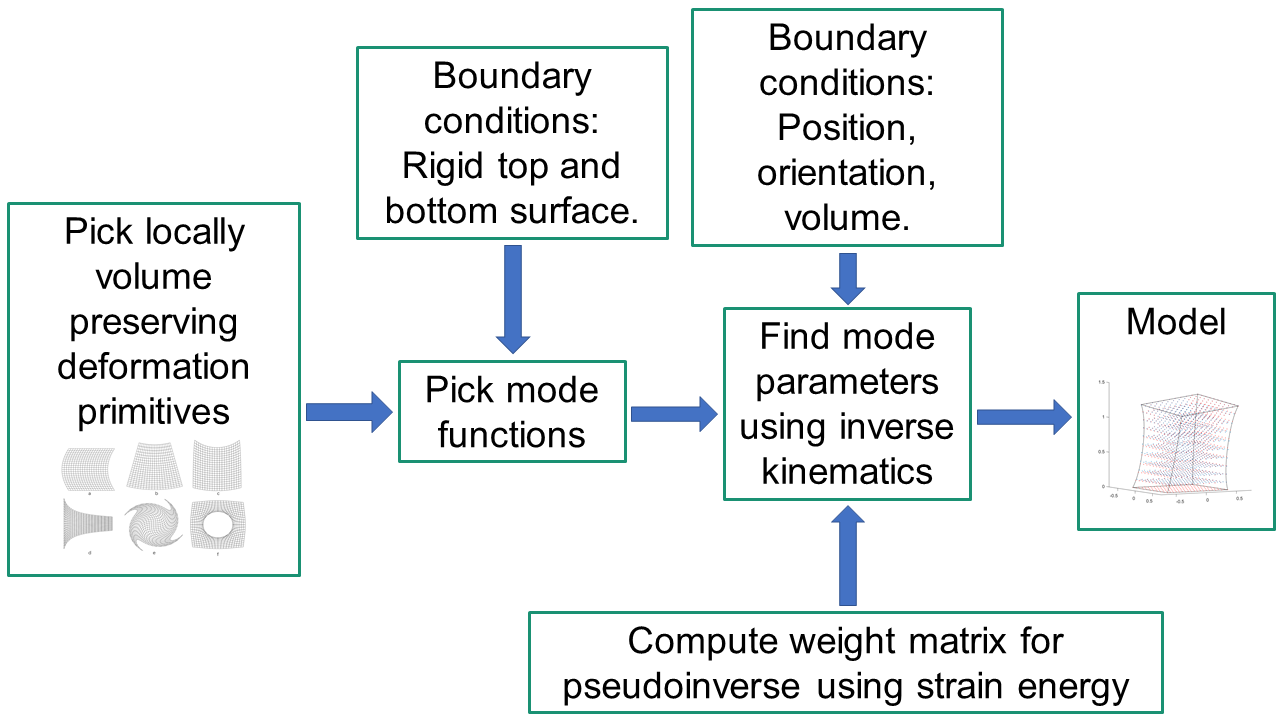}
\caption{The process of modeling.}
\label{process}
\end{figure}

The process of the modeling in this paper is demonstrated in Figure \ref{process}. Details are discussed in this section.
\subsection{Locally volume preserving deformation primitives}
All the modeling works are based on locally volume preserving deformations. A deformation {\bf f} maps a set of referential Cartesian coordinates ${\bf x}=[x_1,x_2,x_3]^T$ into a new set of coordinates ${\bf f}={\bf f}({\bf x})=[f_1,f_2,f_3]^T$. $T$ denotes the transpose of a vector or matrix. The deformation gradient of this deformation is:
\begin{equation}
\nabla_{\bf x} {\bf f} =
\left(\begin{array}{ccc}
\dfrac{\partial f_1}{\partial x_1} & \dfrac{\partial f_1}{\partial x_2} & \dfrac{\partial f_1}{\partial x_3} \\[10pt]
\dfrac{\partial f_2}{\partial x_1} & \dfrac{\partial f_2}{\partial x_2} & \dfrac{\partial f_2}{\partial x_3} \\[10pt]
\dfrac{\partial f_3}{\partial x_1} & \dfrac{\partial f_3}{\partial x_2} & \dfrac{\partial f_3}{\partial x_3} \\
\end{array}\right).
\label{gradient}
\end{equation}
For locally volume preserving deformation,
\begin{equation}
det(\nabla_{\bf x} {\bf f} ) = 1
\label{condition}
\end{equation}
Given two locally volume preserving deformation ${\bf f}({\bf x})$ and ${\bf g}({\bf x})$, the determinant of the gradient of composition {\bf f}({\bf g}({\bf x})) is:
\begin{equation}
det(\nabla_{\bf x}{\bf f}({\bf g}({\bf x})))=det(\nabla_{\bf y}{\bf f}({\bf y}))det(\nabla_{\bf x}{\bf g}({\bf x}))=1\cdot1=1
\label{composition}
\end{equation}
Therefore, the compositions of locally volume-preserving deformations are also locally volume-preserving.

Soft materials are considered incompressible, (\ref{condition}) is hence usually considered as a constraint when analyzing the deflections of soft materials. The advantage of using locally volume preserve deformation to model soft robots is that the constraint of (\ref{condition}) is implicitly included in the functions, so there is no need to apply this constraint again.

As shown in Figure \ref{lvpd}, Chirikjian \cite{chirikjian1995closed} introduced locally volume-preserving primitives such as elongation, twist, shear, bending, source, etc. as a mechanical design tool. Each primitive involves basic scaler functions that define the shape of the deformations (e.g. area of void in source deformation, bending angle in bending deformation, et al.). The basic scaler functions consist of a series of weighted mode functions and can be expressed in the modal form:
\begin{equation}
m(x) = \sum^{M}_{i=1} p_i \phi_i(x),
\label{modalform}
\end{equation}
where $\phi_i(x)$ are the mode functions, $p_i$ are the weight parameters, and $M$ is the number of modes. For example:
\begin{equation}
m(x) = p_1\cdot 1 + p_2 sin(\pi x) + p_3 cos(\pi x)+ p_4 sin(2\pi x),
\label{modes}
\end{equation}
$1, sin(\pi x), cos(\pi x), sin(2\pi x)$ are the mode functions. Once choosen, the functions will be fixed for the model. $p_1,p_2,p_3,p_4$ are the free parameters to be updated. By choosing appropriate mode functions and their free parameters, the shape of the model can be controlled to match the experimental measurement of deformed shape. With sufficient primitive types and modes, an infinite variety of locally volume preserving deformations can be written in closed form by repeated composition.

The primitives used in this paper are introduced below:

\paragraph{Stretching and compression} When stretch or compress a block in the $z$ direction, the cube will elongate or shorten in the $z$ direction, and shrink or expand in the $x$-$y$ plane. If the material is homogeneous, the deformation can be modeled using the nonuniform elongation primitive, ${\bf e(x)}$.
\begin{equation}
\bf{e}(\bf{x}) \,=\, \,
\left(\begin{array}{ccc}
\frac{1}{\sqrt{m_e'(x_3)}} x_1 \\ \\
\frac{1}{\sqrt{m_e'(x_3)}} x_2 \\ \\
m_e(x_3)
\end{array}\right)
\end{equation}
$m_e'(x_3)$ can be expressed as a weighted sum of the mode functions and reflects the elongation rate at the plane $z=x_3$.

\paragraph{Twist}
In this primitive, a 3D block is considered as a stack of parallel 2D planes. As the area of each plane is unchanged, the volume of the block is thus preserved after the deformation. Assuming each plane rotates about the $z$ axis by angle $m_\theta(x_3)$, the deformation can be written as:
\begin{equation}
\bf{t}(\bf{x}) \,=\, \,
\left(\begin{array}{ccc}
x_1 \cos m_\theta(x_3) - x_2 \sin m_\theta(x_3) \\ \\
x_1 \sin m_\theta(x_3) + x_2 \cos m_\theta(x_3)\\ \\
x_3
\end{array}\right)
\end{equation}
$m_\theta(x_3)$ determines the rotation angle at each horizontal plane and can be chosen to be a weighted sum of modes.

\paragraph{Shear}
A shear deformation ${\bf s(x)}$ is defined as below, horizontal planes move along the $x$ and $y$ axis while no deformation exists in the $z$ direction:
\begin{equation}
\bf{s}(\bf{x}) \,=\, \,
\left(\begin{array}{ccc}
x_1 + m_{s1}(x_3)\\ \\
x_2 + m_{s2}(x_3)\\ \\
x_3
\end{array}\right)
\end{equation}
$m_{s1}(x_3), m_{s2}(x_3)$ are the scaler functions that can be described using modes.

\paragraph{2D Bend}
Bending deformation is defined based on the backbone curve, which captures the configuration of a continuum.
\begin{equation}
\begin{array}{ccc}
{\bf b}_2({\bf x}) &=& {\bf a}(x_3) + r(x_2,x_3) {\bf n}(x_3) \\ \\
&=& {\bf a}(x_3) + \frac{1-\sqrt{1-2 m_\kappa(x_3) x_2}}{m_\kappa(x_3)} {\bf n}(x_3)
\end{array}
\end{equation}
${\bf a}(x_3)$ is the 2D backbone curve, while $m_\kappa(x_3)$ is the curvature of the backbone curve parameterized by its arclength $x_3$. In order to avoid singularities, the constraint $\frac{1}{2}>m_\kappa x_2$ is applied. The backbone is initially located in the $y$-$z$ plane. To add more freedom and enable bending in 3D place, the backbone curve can be rotated about the $z$ axis.

\paragraph{3D Bend}
A 2D bending deformation can be extended to 3D space. Given a 3D bending deformation of this form:
\begin{equation}
{\bf b}_3({\bf x}) = {\bf a}(x_3) + \nu(x_1,x_2,x_3) {\bf n}(x_3) + \beta(x_1,x_2,x_3) {\bf b}(x_3) \,,
\label{bendtwist}
\end{equation}
${\bf a}(x_3)$ is the 3D backbone curve, ${\bf n}$ is the normal vector and ${\bf b}$ is the binormal vector. The detailed derivation is demonstrated in Appendix A. There are many different choices of $\nu$ and $\beta$ that will result in local volume preservation. A simple one is defined by letting $\beta=x_1$ and $\nu=\frac{1 - \sqrt{1 - 2 \kappa x_2}}{ \kappa}$, where $\kappa(x_3)$ is the curvature of the backbone curve. Since the Frenet frames twist along the tangent as they traverse the curve, this bending deformation should be composed with a twist deformation along the $x_3$ axis to form a minimally varying frame. Assume the twist deformation rotates the frame about the $x_3$ axis by angle $\theta (x_3)$. The new frame has globally minimal twist if \cite{chirikjian2016harmonic}:
\begin{equation}
\theta_1(s)=-\int_{0}^{s} \tau(\sigma) \,d\sigma,
\end{equation}
where $\tau(s)$ is the torsion of the curve, and $s$ is the arclength of the curve. In order to preserve the orientation of the bottom plane, let:
\begin{equation}
\theta(x_3)=\theta_1(x_3)+\theta_2,
\end{equation}
where $\theta_2$ is a constant equal to the angle between the normal vector of the backbone curve at $x_3=0$ and the $x_2$ axis.

\paragraph{Source}
The source deformation ${\bf c(x)}$ can simulate cavity generation in material. This deformation can model the bulge of chambers made of soft materials after inflation.
\begin{equation}
\bf{c}(\bf{x}) \,=\, \,
\left(\begin{array}{ccc}
(m_c(x_3)+x_1^2+x_2^2)^{\frac{1}{2}} \frac{x_1}{(x_1^2+x_2^2)^{\frac{1}{2}}}\\ \\
(m_c(x_3)+x_1^2+x_2^2)^{\frac{1}{2}} \frac{x_2}{(x_1^2+x_2^2)^{\frac{1}{2}}} \\ \\
x_3
\end{array}\right)
\end{equation}
$m_c(x_3)$ is the function expandable in modes. It reflects the radial change of the void within the material.

\subsection{Boundary conditions}
A common approach to increase the DOFs of soft robots is to connect a series of sections with different functionality. As the connecting structures of the individual sections are usually rigid, it is useful to analyze models with rigid surface boundary conditions applied on both top and bottom.

To incorporate boundary conditions with primitives, we need to arrange the primitives in proper order and select appropriate modes for scaler functions. Commutable primitives such as stretch, twist, shear, and source do not affect the parallelism between horizontal planes, which means the order of compositions is irrelevant. However, bending deformation yields unparallel horizontal planes, which has to be composed last.

After obtaining the appropriate order of primitives, the modes are then constrained based on the boundary conditions applied. Table \ref{boundaryconditiontable} formulates the rigid surface boundary conditions applied to the modes introduced above. The mode functions should be tailored to satisfy the boundary conditions automatically. The specific choice of modes is introduced in the later section.
Although the boundary condition can also be achieved by adjusting suitable free parameters, choosing the modal basis functions that automatically satisfy the condition reduces the tuning work required for the parameters.

\begin{table}[ht]
\caption{The boundary conditions for elongation, shear, twist, bending, and source primitives. $h$ stands for the height of the object.}
\begin{center}
\begin{tabular}{c c c}
\hline
Primitive & Boundary condition & Explanation \\
\hline
Elongation & $\frac{1}{\sqrt{m_e'(0)}} = \frac{1}{\sqrt{m_e'(h)}} = 1$ & \makecell[c]{Limit elongation rate at top and bottom \\plane to 1} \\
Shear & $m_{s1}(0)= m_{s2}(0)=0$ & Fix the bottom plane\\
Twist & $m_\theta(0)=0$ & Fix the bottom plane\\
Bending & $m_\kappa(0)=m_\kappa(h)=0$ & \makecell[c]{Preserve the shape of top and bottom \\plane by making the curvature zero}\\
Source & $m_c(0)=m_c(h)=0$ & \makecell[c]{Set strength of the source to 0 to \\preserve top and bottom plane}\\
\hline
\end{tabular}
\end{center}
\label{boundaryconditiontable}
\end{table}


\subsection{Kinematics}
After acquiring the composition of primitives, the shape of the model can be controlled by manipulating the mode parameters. Based on observation, when only kinematical boundary conditions are applied, the deformation is not sensitive to material property. The configuration of the deformed block can then be obtained using inverse kinematics without strain energy. The simplest inverse kinematics based on the Jacobian matrix is described here:

Denote the state of interest as ${\bf d}$, which could be the volume of inflation, the position of the top plane, etc, and the vector that contains all the weight parameters in the kinematic deformation as ${\bf p}$. In every step, current state ${\bf d}({\bf p})$ is assumed to move a small step $\Delta {\bf d}$ towards desired state ${\bf d}_d$, where the deformation of the object is studied. $J$ is the Jacobian matrix, where $J_{ij}=\frac{\partial d_i}{\partial p_j}$. The change of parameters ${\bf p}$ should be:
\begin{equation}
\Delta {\bf p} = J^{-1}\Delta {\bf d}.
\label{inflateupdate}
\end{equation}
If the number of free parameters is larger than the number of boundary conditions, the right Jacobian pseudoinverse is applied.
\begin{equation}
J^+=W^{-1} J^T (J W^{-1} J^T)^{-1}
\label{jpi}
\end{equation}
$W$ is the weight matrix. $W$ can be set to the identity matrix, which generates the result where all the parameters have the same impact. $W$ can also be tuned using mechanics to closely describe the deformation, which will be discussed in the next subsection.
${\bf p}$ is updated until ${\bf d}_d$ is reached. At this point, the suitable mode parameters are found to present the deformation that satisfies the state of interest.

Another method is the projected gradient method. This method involves moving along the negative gradient of the cost function to minimize cost \cite{chirikjian1995kinematically}. Let ${Z}_k={\mathbb{I}}-J^+ J$. The columns of ${ Z}_k$ are the basis for the null space of $J$. The mode parameters ${\bf p}$ are iteratively updated to satisfy the boundary conditions using the following equation.
\begin{equation}
{\bf p}_{k+1}={\bf p}_k+J^+ ({\bf d}_d - {\bf d}({\bf p}_k)) - \alpha {Z}_k \nabla G({\bf p}_k)
\label{Jpi}
\end{equation}
$\alpha$ is a positive constant smaller than 1. $G({\bf p}_k)= {\bf p}_k^T {\bf p}_k$ is the configuration cost.

This method can handle simple boundary conditions, but there are better options when 3D rotation is involved. The following method by Kim et al.\cite{kim2006conformational} could satisfy desired orientation without the usage of Euler angles, hence avoiding the singularity problem.
The position and orientation are described using elements in group SE(3), the group of rigid-body displacements. Given $g \in SE(3)$,
\begin{equation}
g \,=\,
\left(\begin{array}{ccc}
R & {\bf t}\\ \\
{\bf 0}^T & 1
\end{array}\right)
\end{equation}
where $R$ is a $3 \times 3$ rotation matrix that describes the orientation of the object. It is an element of SO(3), the special orthogonal group that describes the orientation and rotation in 3D space. ${\bf t}$ is a $3 \times 1$ vector describes the position or translation. The group operation of SE(3) and SO(3) is matrix multiplication.

Define the ${ }^{\vee}$ operator moves the elements of SO(3) to the real vector space of dimension 3. In this case,
\begin{equation}
{\bf t}=\left(\sum_{i=1}^{3} t_i E_i \right)^{\vee}=\sum_{i=1}^{3} t_i E_i^{\vee} ,
\end{equation}
where
\begin{equation}
E_1 \,=\,
\left(\begin{array}{ccc}
0 & 0 & 0\\ \\
0 & 0 & -1\\ \\
0 & 1 & 0
\end{array}\right);
E_2 \,=\,
\left(\begin{array}{ccc}
0 & 0 & 1\\ \\
0 & 0 & 0\\ \\
-1 & 0 & 0
\end{array}\right);
E_3 \,=\,
\left(\begin{array}{ccc}
0 & -1 & 0\\ \\
1 & 0 & 0\\ \\
0 & 0 & 0
\end{array}\right).
\end{equation}
$E_i$ are the basis matrices for the Lie algebra so(3), i.e. $E_i^\vee = {\bf e}_i$. Another operator associated with ${ }^{\vee}$ is the $\hat{ }$ operator, which moves the vector space backward, i.e. $(\hat{\bf t})^\vee = {\bf t} \in \mathbb{R}^3$.

For SE(3), there are similar operators. The $\vee$ operator for SE(3) is denoted as $\dvee$ and defined as
\begin{equation}
\boldsymbol{\xi} =\left(\sum_{i=1}^{6} \xi_i \tilde{E}_i\right)^{\dvee}=\sum_{i=1}^{6} \xi_i \tilde{E}_i^{\dvee},
\end{equation}
where
\begin{equation}
\begin{aligned}
\tilde{E}_1 &\,=\,
\left(\begin{array}{cccc}
0 & 0 & 0 & 0\\ \\
0 & 0 & -1 & 0\\ \\
0 & 1 & 0 & 0\\ \\
0 & 0 & 0 & 0
\end{array}\right);
\tilde{E}_2 \,=\,
\left(\begin{array}{cccc}
0 & 0 & 1 & 0\\ \\
0 & 0 & 0 & 0\\ \\
-1 & 0 & 0 & 0\\ \\
0 & 0 & 0 & 0
\end{array}\right);
\\
\tilde{E}_3 &\,=\,
\left(\begin{array}{cccc}
0 & -1 & 0 & 0\\ \\
1 & 0 & 0 & 0\\ \\
0 & 0 & 0 & 0\\ \\
0 & 0 & 0 & 0
\end{array}\right);
\tilde{E}_4 \,=\,
\left(\begin{array}{cccc}
0 & 0 & 0 & 1\\ \\
0 & 0 & 0 & 0\\ \\
0 & 0 & 0 & 0\\ \\
0 & 0 & 0 & 0
\end{array}\right);
\\
\tilde{E}_5 &\,=\,
\left(\begin{array}{cccc}
0 & 0 & 0 & 0\\ \\
0 & 0 & 0 & 1\\ \\
0 & 0 & 0 & 0\\ \\
0 & 0 & 0 & 0
\end{array}\right);
\tilde{E}_6 \,=\,
\left(\begin{array}{cccc}
0 & 0 & 0 & 0\\ \\
0 & 0 & 0 & 0\\ \\
0 & 0 & 0 & 1\\ \\
0 & 0 & 0 & 0
\end{array}\right).
\end{aligned}
\end{equation}
$\tilde{E}_i$ are the basis matrices for the Lie algebra se(3), and the $\hat{\vphantom{\rule{1pt}{5.5pt}}\smash{\hat{ }}}$ operator is defined as $(\hat{\vphantom{\rule{1pt}{5.5pt}}\smash{\hat{{ \boldsymbol{\xi} } }}})^{\dvee}=\boldsymbol{\xi} \in \mathbb{R}^6$. From these definition, it is clear that $E_i \leftrightarrow {\bf e}_i$ and $\tilde{E}_i \leftrightarrow \tilde {\bf {e}}_i$, the natural unit basis elements for $\mathbb{R}^3$ and $\mathbb{R}^6$, respectively.

Define the adjoint operator $Ad$ as
\begin{equation}
Ad(g) \,=\,
\left(\begin{array}{ccc}
R & 0_{3\times 3}\\ \\
\hat{\bf t}R & R
\end{array}\right).
\end{equation}

The 3D backbone curve is considered as an inextensible fiber. The orientation of the curve $A(s)$ can be computed from the angular velocity $ \boldsymbol{\omega} (s)$ using:
\begin{equation}
\frac{dR}{ds}=R\left(\sum^{3}_{i=1} \omega_i(s)E_i \right).
\end{equation}
Then the configuration can be computed using:
\begin{equation}
{\bf t}(s)=\int^{s}_{0} R(\delta){\bf e}_3 d\delta,
\end{equation}
where ${\bf t}(s)$ is the position of the point on backbone curve at arclength $s$. Assume the total length of the backbone curve is $L$, then ${\bf t}(L)$ and $R(L)$ is the desired position and orientation.
The positional and orientational boundary conditions and inextensible constraint is achieved by solving: 
\begin{equation}
\frac{d \boldsymbol{\omega}}{ds}=
\left(
\begin{array}{c}
-\boldsymbol{\lambda}^T R {\bf e}_2 \\
\boldsymbol{\lambda}^T R {\bf e}_1 \\
0
\end{array}
\right).
\end{equation}
No modes are involved in this method, $\bf p$ are the free parameters that need to be determined. $[p_1, p_2, p_3]^T=\boldsymbol{\omega}$ is the angular velocity of the motion $g(t)$, and $[p_4, p_5, p_6]^T=\boldsymbol{\lambda}$ is the Lagrange multiplier to enforce the boundary conditions. 

\begin{figure}[ht]
\centering
\includegraphics[scale=0.2]{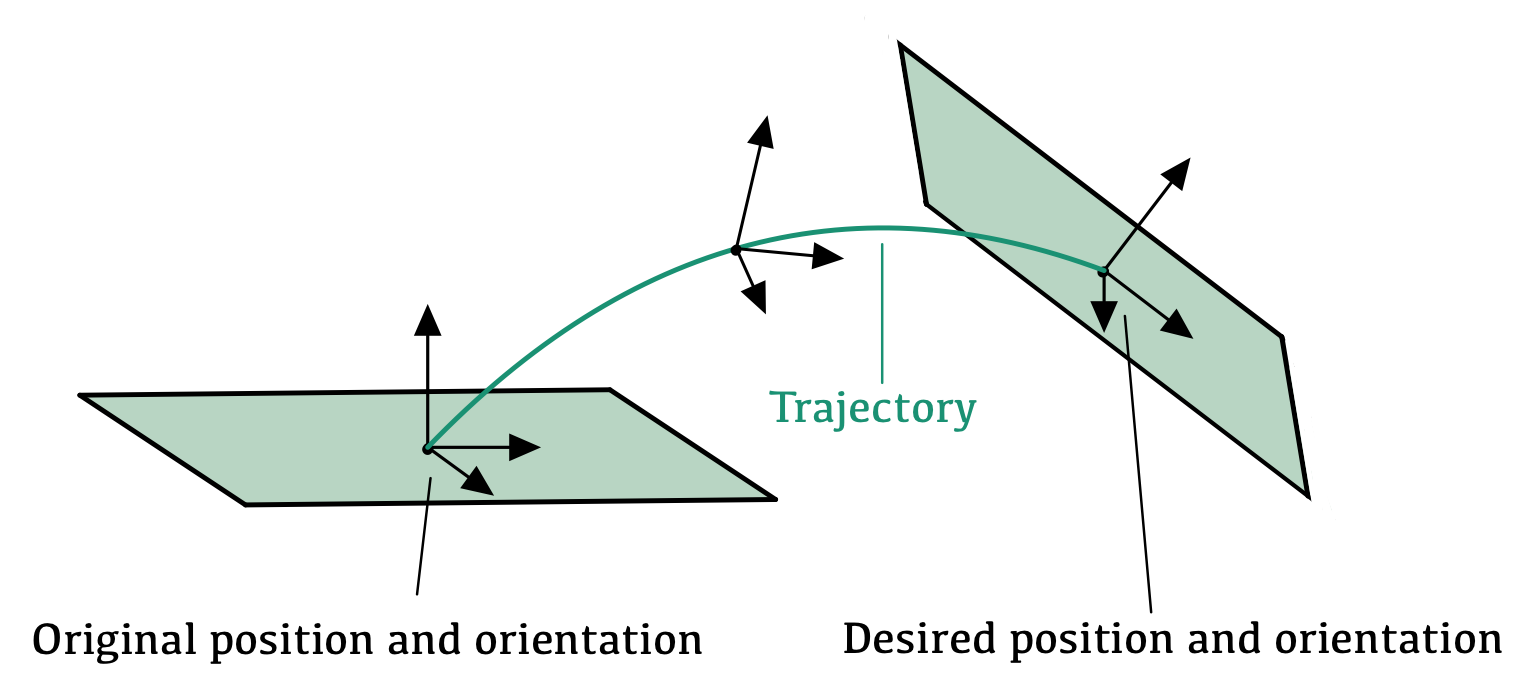}
\caption{Trajectory of the end effector while updating the parameters}
\label{traj}
\end{figure}

Denote the homogeneous transformation matrix of the defined trajectory as $g_{p}$. This method defines a path from original position and orientation $g_p(0)$ to desired ones $g_p(1)$, as shown in Figure \ref{traj}.

The velocity condition is applied first:
\begin{equation}
\left( g_{p}^{-1}\dfrac {dg_{p}}{dt}\right) ^{\dvee}=J \left( {\bf p} \right) \dfrac {d{\bf p} }{dt}
\end{equation}
\begin{equation}
\dfrac {d{\bf p}_{k}}{dt}=J^{+} Ad\left( g^{-1}g_{p}\left( t_{k}\right) \right) \left( g^{-1}_{p}\left( t_{k}\right) \dfrac {dg_{p}}{dt}\left( t_{k}\right) \right) ^{\dvee}
\end{equation}
The Jacobian matrix is calculated using centered finite difference approximation. 

There is also a position correction term for the update of parameters:
\begin{equation}
{\bf p}^{c}_{k}=J^{+} Ad\left( g^{-1}g_{p}\left( t_{k}\right) \right) \left( \log \left( g^{-1}\left( t_{k}\right) g_{p}\left( t_{k}\right) \right) \right) ^{\dvee}
\end{equation}
$\log$ is the matrix logarithm. Iteratively update $\bf p$ to get the desired configuration.
\begin{equation}
{\bf p}_{k+1}={\bf p}_{k}+\Delta t\dfrac {d{\bf p}_{k}}{dt}+{\bf p}^{c}_{k}
\end{equation}
When using the kinematics method to compute the mode parameters, we must set the original parameters to values of undeformed configuration. In this way, the configuration with low strain energy can be obtained by kinematics without constitutive modeling.

\subsection{Mechanics}
When redundant DOFs are available in the system, using minimization of strain energy to guide the update of parameters can improve the performance of the model.
Given a comoposition of a series of primitives, ${\bf f}= {\bf f}_n({\bf f}_{n-1}(\ldots{\bf f}_1({\bf x})\ldots))$, vector ${\bf p}$ contains all the free parameters in ${\bf f}$. {\bf F} is the deformation gradient. ${\bf B}={\bf F}{\bf F}^T$ is symmetric and positive definite tensor of left Cauchy deformation. The strain energy function of Mooney-Rivlin model is expressed as a function of {\bf B} through its principle invarians $I_{\bf B}\doteq\
\begin{Bmatrix}
I_1({\bf B}),I_2({\bf B}),I_3({\bf B})
\end{Bmatrix}
,$
where $I_{\bf B}$ is defined as:
\begin{equation}
\begin{aligned}
I_1({\bf B})&=tr({\bf B})\\
I_2({\bf B})&=\frac{1}{2}[tr^2({\bf B})-tr({\bf B}^2)]\\
I_3({\bf B})&=det({\bf B})
\end{aligned}
\label{invarians}
\end{equation}
These invariants are essentially kinematic quantities. Our emphasis is on pulling $I_3$ out of the realm of constitutive modeling, and handling it as kinematics. However, classically in mechanics, constitutive laws are built on all three invariants. For two parameters model, the strain energy density function is given as:
\begin{equation}
\psi = C_{10} (I_1({\bf B})-3)+C_{01}(I_2({\bf B})-3)+\frac{1}{d}(J-1)K
\label{mr1}
\end{equation}
where $J^2=I_3({\bf B})$. For incompressible materials, $I_3({\bf B})=1$, the strain energy density function becomes:
\begin{equation}
\psi = C_{10} (I_1({\bf B})-3)+C_{01}(I_2({\bf B})-3)
\label{mr}
\end{equation}
$C_{10}$ and $C_{01}$ are two constant coefficients determined experimentally for any specific material. When $C_{01}=0$, the Mooney-Rivlin model becomes the Neo-Hookean model. To calculate the total energy of the whole body, $\psi$ is integrated over the entire volume. The strain energy $E_e$:
\begin{equation}
E_e({\bf{p}})=\iiint_{V}\psi \, dx\,dy\,dz
\label{elastomerenergy}
\end{equation}
In cylindrical coordinate system, the total strain energy can be calculated as:
\begin{equation}
E_e({\bf{p}})=\iiint_{V}\psi \, dx\,dy\,dz=\iiint_{V}\psi \, r dr\,d\theta \,dh
\end{equation}
This integral is computed numerically. Assuming the system is divide into $N$ small volumes $V_i$,
the integration becomes the sum of the product of the strain energy density at the corresponding sample point and its corresponding volume.
\begin{equation}
E_e({\bf{p}})=\sum_{i=1}^{N} V_i \psi_i
\label{strainenergy}
\end{equation}

The strain energy computed by (\ref{strainenergy}) is then used to compute the weight matrix $W$ in the Jacobian pseudoinverse in (\ref{jpi}). The pseudoinverse mininizes the quadratic cost $c$:
\begin{equation}
c={\bf p}^T W {\bf p}
\end{equation}
The following steps can compute the weight matrix W that minimizes the strain energy when deformation is very small. For large deformations, it also helps lower the strain energy of the configuration.

Choose $n$ sets of random small values: ${\bf p}_i, {\bf p}_2, ..., {\bf p}_n$. $m$ is the length of ${\bf p}$. $c_i$ is the strain energy of the structure when ${\bf p}_i$ is used. Then we have:
\begin{equation}
c_i=\sum_{k=1}^{\frac{m(m+1)}{2}} {\bf p}_i^T M_k {\bf p}_i w_k
\label{66}
\end{equation}
$M_k$ is a symmetric matrix that has either one entry on the diagonal or two entries on both sides of the diagonal that equal one. The other entries are zero. For example, when $m=2$, there should be $\frac{m(m+1)}{2}=3$ different $M$ matrices:
\begin{equation}
M_1=
\left(
\begin{array}{cc}
1&0 \\
0&0 \\
\end{array}
\right)
,
M_2=
\left(
\begin{array}{cc}
0&0 \\
0&1 \\
\end{array}
\right)
,
M_3=
\left(
\begin{array}{cc}
0&1 \\
1&0 \\
\end{array}
\right)
\end{equation}
Equation \ref{66} can be written as:
\begin{equation}
\sum_{k=1}^{\frac{m(m+1)}{2}} a_{ki} w_k = c_i
\label{66a}
\end{equation}
$a_{ki}$ is the entry $ki$ of matrix $A$. The above equation can be rewritten as follows:
\begin{equation}
A{\bf w}={\bf c}
\end{equation}
${\bf c}$ contains value from $c_1$ to $c_n$. Use unweighted pseudoinverse to solve for ${\bf w}$:
\begin{equation}
{\bf w}=A^+{\bf c}
\end{equation}
${\bf w}$ contains all the entries of weight matrix $W$. Using this weight matrix in the pseudoinverse, the result will converge to a result that has lower total strain energy compared to unweighted ones.

\section{Implementation and comparison with FEM and physical experiments}
In this session, our modeling technique is applied to three different objects: a rubber chamber, a slender cylindrical rubber rod, and a rubber block. As pneumatic actuators play an important role in soft robotics, the first model of interest is the inflation of a cylindrical chamber. We also model the deformation of a soft rod as it is a crucial configuration of the continuum robots and rod-driven robots. To further demonstrate the capability of our method with complex boundary conditions, a rubber block is modeled to study the variation of 6 DoFs on the top plane.

\begin{figure}[ht]
\centering
\includegraphics[scale=0.2]{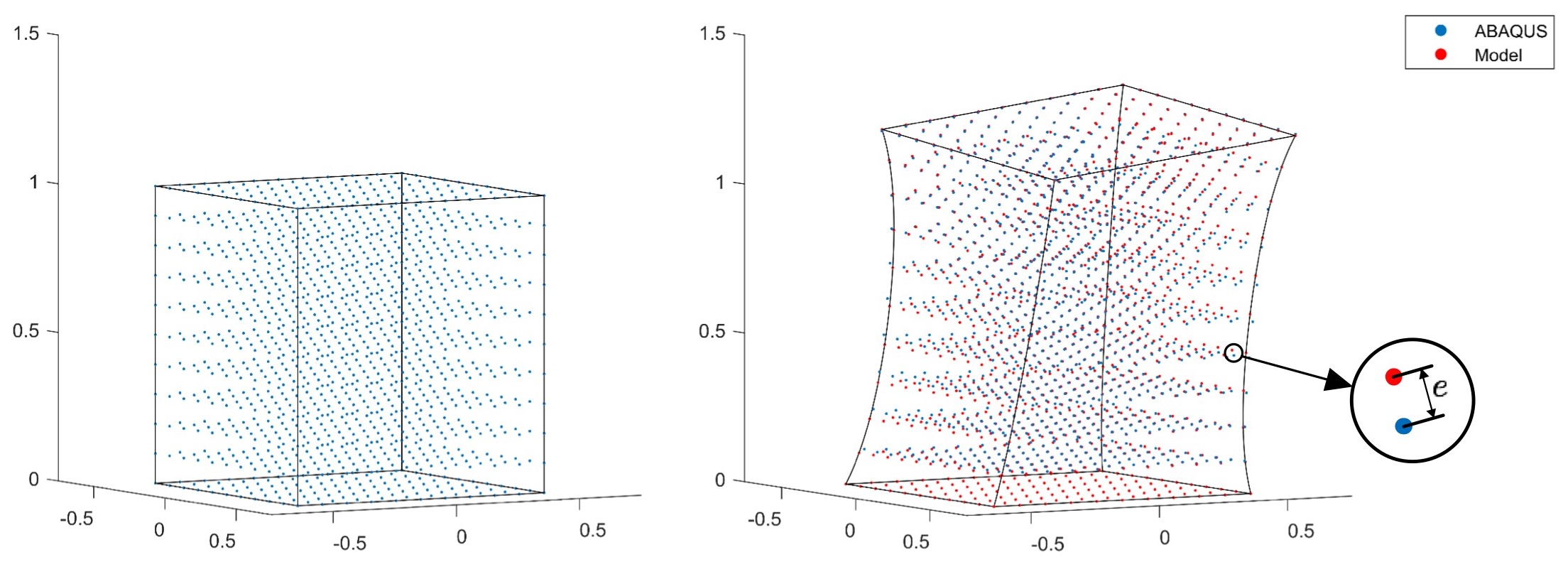}
\caption{The nodes within the system, and the error between a pair of nodes.}
\label{error}
\end{figure}

To demonstrate the accuracy of our method, our results are compared with ABAQUS and physical experiments. ABAQUS is a software for finite element analysis. It devides the domain into simple polyhedral elements and uses displacements at the vertices as DOFs to simulate the deformation of the system. At the same time. the coordinate of these nodes are also fed into our model. The output result of our method and ABAQUS can then be compared at the corresponding node.
Assuming $l$ nodes are sampled to represent the system. ${\bf o}_i({\bf p})$ is the original position of the $i$th node. ${\bf f}_i({\bf p})$ is the position after the deformation computed by primitives, and ${\bf a}_i({\bf p})$ is the position computed by ABAQUS. The average error $E$ is calculated as the squared root of the sum of the squared error normalized by the total displacement.
\begin{equation}
E=\left(\frac{\sum_{i=1}^{l} e_i^2}{\sum_{i=1}^{l} d_i^2}\right)^{\frac{1}{2}}=\left(\frac {\sum_{i=1}^{l} \Vert{\bf a}_i({\bf p})-{\bf f}_i({\bf p})\Vert^2}{\sum_{i=1}^{l} \Vert{\bf a}_i({\bf p})-{\bf o}_i({\bf p})\Vert^2}\right)^{\frac{1}{2}}
\label{abacom}
\end{equation}
$d_i$ represents the displacement of node $i$. $e_i$ represents the distance difference of node $i$ by two different methods, as illustrated in Figure \ref{error}.

Our method is also benchmarked against measurements from the physical experiment. Most of the soft objects in this session are cast using Ecoflex 00-30. An exception is the rod used for testing the model for 3D bending. It is made of Dragonskin 30, which is stiffer than Ecoflex 00-30 and allows us to neglect the impact of gravity during the experiment. The parameters of Mooney-Rivlin model for Ecoflex 00-30 are: $C_{10}=5.6 kPa, C_{01}=6.3 kPa$, for Dragonskin 30 are: $C_{10}=1.19 kPa, C_{01}=23.028 kPa$. All the objects have markers drawn on the surface for coordinate positioning. The configurations of soft objects are captured by Zed 2 stereo cameras. Zed takes 2 images from different angles at the same time to allow the stereo matching, and hence 3D coordinates calculation. Zed cameras are calibrated using Zhang's method \cite{zhang2000flexible}, which produces the intrinsic matrix of the camera. The undistorted version of the stereo images can then be obtained and fed into MATLAB computer vision toolbox, which offers a triangulation tool for 3D coordinates positioning.

\subsection{Modeling with boundary conditions: chamber}
\begin{figure}[ht]
\centering
\includegraphics[scale=0.15]{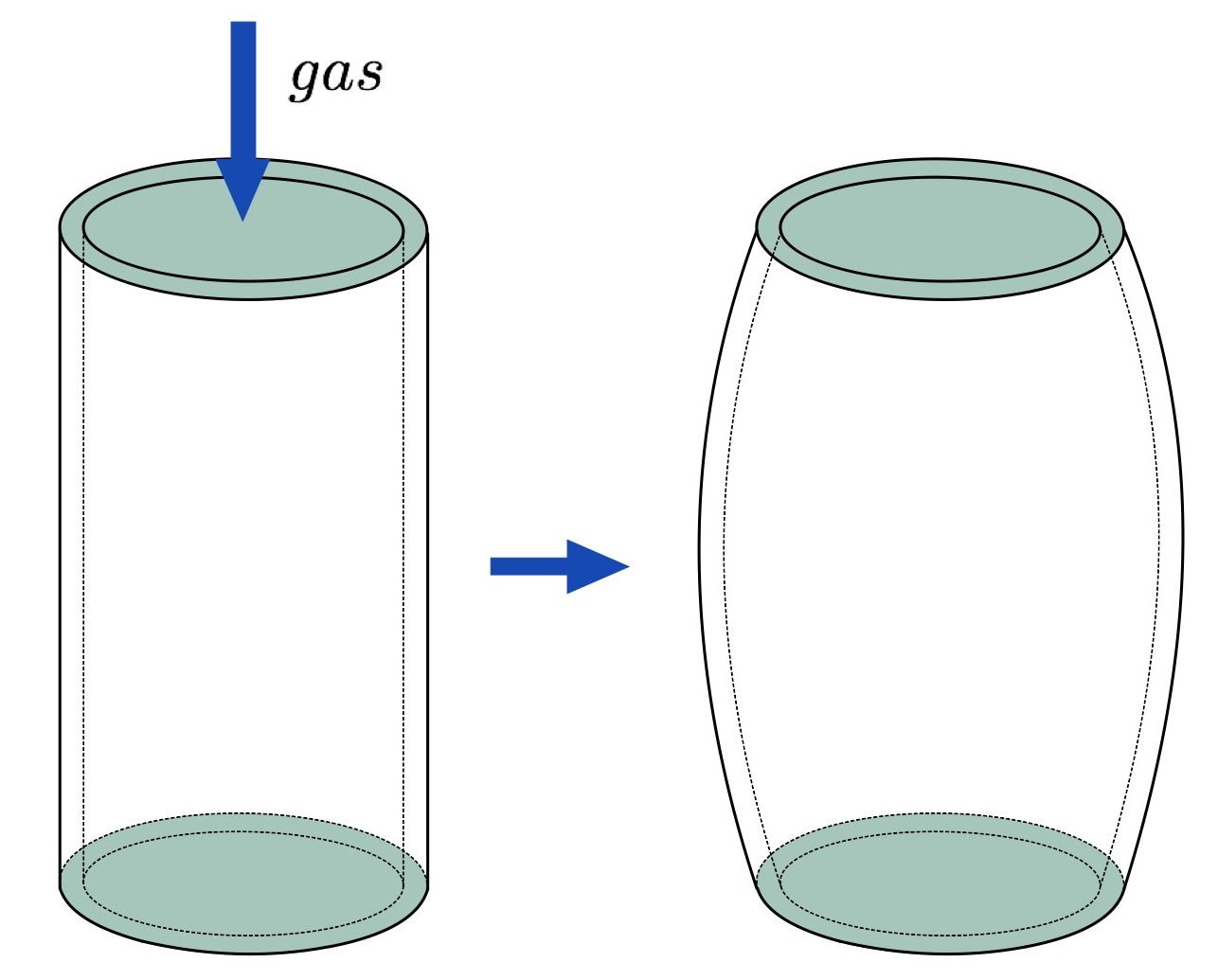}
\caption{Rubber chamber before and after inflation. The green surfaces are rigid.}
\label{inflate}
\end{figure}

As a case study of the method presented in this paper, consider a cylindrical chamber with even wall thickness. The original dimensions of the chamber are 1.6 $cm$ in the radius of chamber crosssection, 0.2 $cm$ in wall thickness, and 10 $cm$ in chamber height. The top and the bottom of the chamber are sealed with rigid material. Upon inflation, the body of the chamber will bulge out, as shown in Figure \ref{inflate}.

This deformation can be modeled using a single source deformation $\bf c(x)$. The strength of the source changes along the $x_3$ axis. Since the chamber has even wall thickness at all heights, it will bulge symmetrically after inflation. To ensure symmetry and also satisfy the boundary conditions in Table \ref{boundaryconditiontable} automatically, the modal expension is chosen as below:
\begin{equation}
\begin{aligned}
m_c(x_3)&=p_1 \sin\left(\frac{\pi x_3}{L}\right)+p_2 \sin\left(\frac{3 \pi x_3}{L}\right)+p_3 \sin\left(\frac{5 \pi x_3}{L}\right)\\
&+p_4 \sin\left(\frac{7 \pi x_3}{L}\right)+p_5 \sin\left(\frac{9 \pi x_3}{L}\right).
\end{aligned}
\end{equation}

\begin{figure}[ht]
\centering
\includegraphics[scale=0.3]{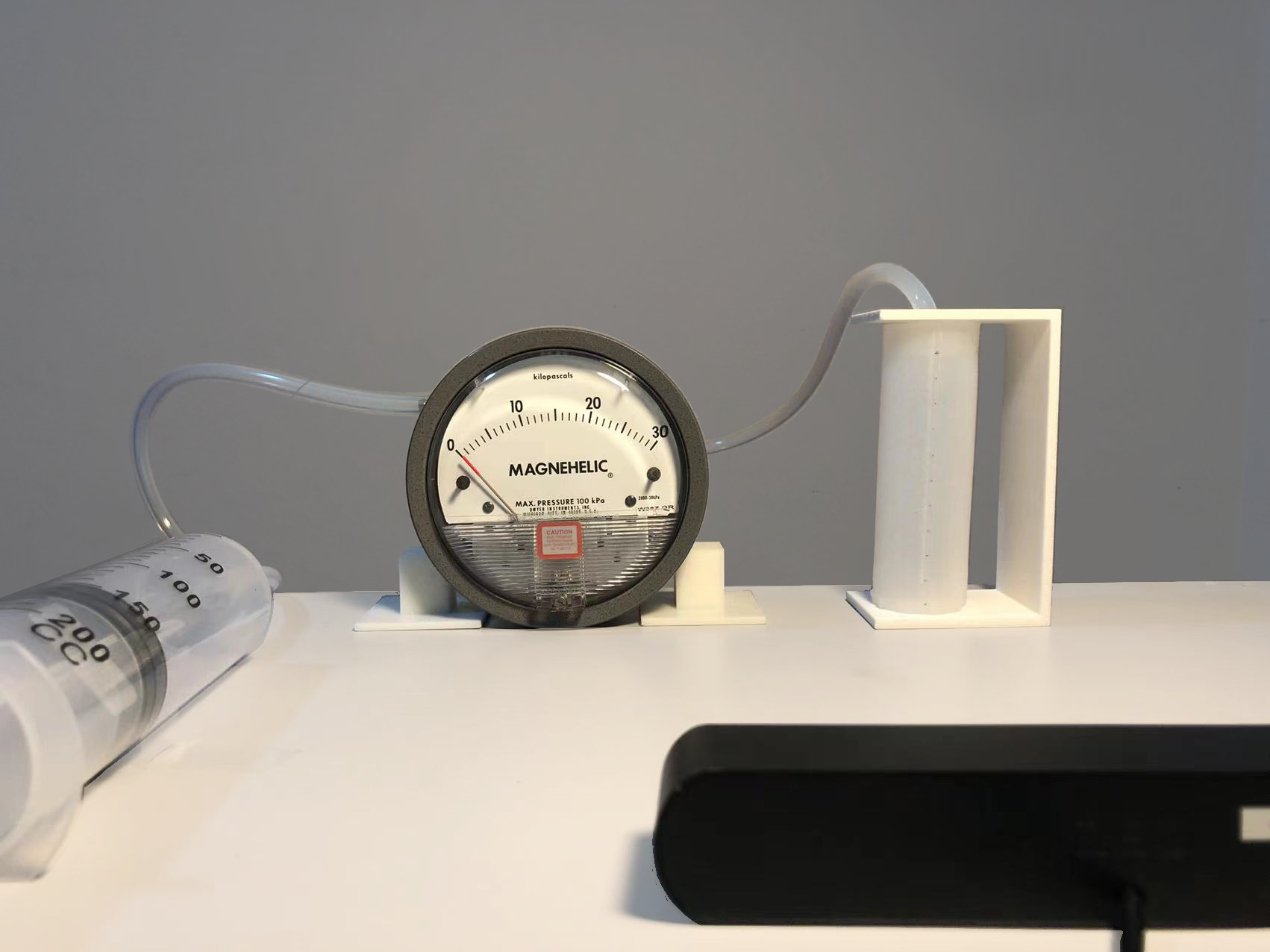}
\caption{Experiment setup of the chamber case}
\label{chambersetup}
\end{figure}

The experiment setup to test the model is shown in Figure \ref{chambersetup}. A rubber chamber is attached to a rigid frame. A 200 $mL$ syringe is connected to the chamber through a silicone tube. The air tube goes through a pressure gauge. The ZED 2 stereo camera for image capturing is placed at the side of the chamber.

The chamber can be inflated by pushing the handle of the syringe. Denote the total volume of air inside the chamber, syringe, tube and gauge before the inflation as $V_b$, atmospheric pressure as $P_b$. Read the inflation volume $V_i$ and pressure $P_a$. Assuming the ideal gas law applied, the total volume of the air in the system after inflation becomes $V_a=V_b \frac{P_b}{P_a}$. Then the volume change of the chamber $V_d$ is:
\begin{equation}
V_d=V_i-(V_b-V_a)
\end{equation}

The boundary condition in ABAQUS is the pressure applied to the inner surface of the chamber while the boundary condition in our method is the volume of the chamber.


\subsection{Modeling with boundary conditions: rod}
\begin{figure}[ht]
\centering
\includegraphics[scale=0.1]{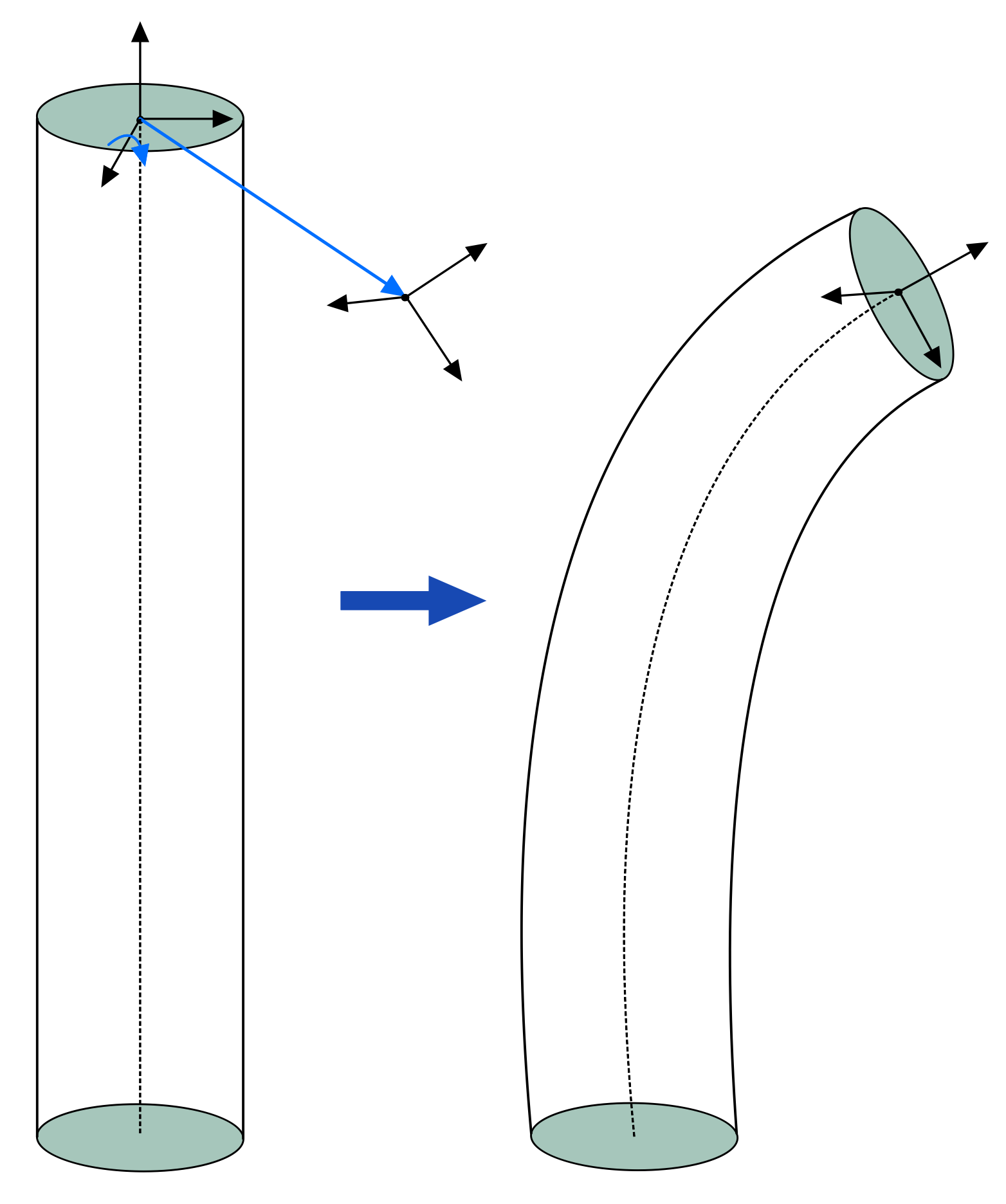}
\caption{Rod with boundary conditions. The dashed line at the center of the cylinder is the backbone curve, whose shape reflects the shape of the rod.}
\label{rod}
\end{figure}

The second case study is a soft rod with a high aspect ratio. In this scenario, bending is the major deformation in the system such that bending primitive ${\bf b}_2 ({\bf x})$ or ${\bf b}_3({\bf x})$ dominates the primitives, and other primitives are negligible. Hence, it is sufficient to only model bending in our method. The key to model the deformation of the rod is to capture the ``backbone" curve of the rod, which is the curve that located at the center of the rod. It can be used to describe the shape of the deformed rod, as shown in Figure \ref{rod}. The dimensions of the undeformed rod are 15 $cm$ of the height, and 0.45 $cm$ of the radius of crosssection.

In the 2D case, the backbone curve can be modeled with curvature scaler function as below.
\begin{equation}
\begin{aligned}
m_{\kappa}(x_3) &=p_1 \sin\left(\frac{\pi x_3}{L}\right)+p_2 \sin\left(\frac{2 \pi x_3}{L}\right)+p_3 \sin\left(\frac{3 \pi x_3}{L}\right)\\
&+p_4 \sin\left(\frac{4 \pi x_3}{L}\right)+p_5 \sin\left(\frac{5 \pi x_3}{L}\right)+p_6 \sin\left(\frac{6 \pi x_3}{L}\right)
\end{aligned}
\end{equation}
These modes are capable of satisfying a wide range of boundary conditions while generating proper configuration. Since the curvature at $x_3=0$ and $x_3=L$ are zero, the top and bottom planes are preserved automatically. In the 3D case, the backbone curve is solved numerically, hence no modes are needed. However, the curvature at $x_3=0$ and $x_3=L$ might not be zero, which leads to the deformation of the plane. For the rod case, this deformation can be neglected since the area of the crosssection is small compared to the height, and the curvature is usually small at these surfaces.

\begin{figure}[ht]
\centering
\includegraphics[scale=0.4]{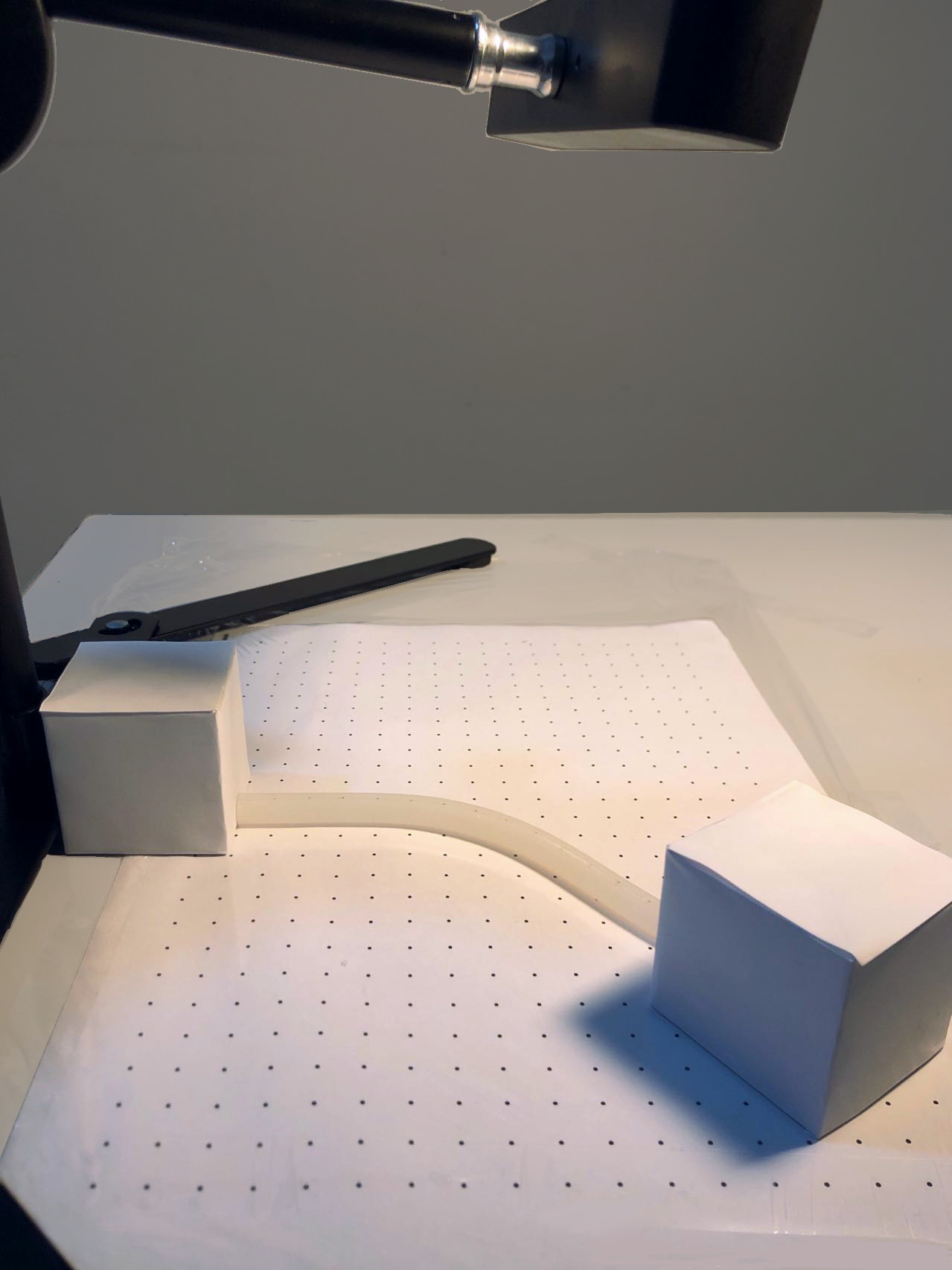}
\caption{Experiment setup of the rod case}
\label{Rodset}
\end{figure}
Figure \ref{Rodset} shows the experiment setup to test the rod model for 2D deformation. A rubber rod is placed on a flat horizontal surface. Lubricant is applied to the surface so that the rod can slide on the surface smoothly. The two ends of the rod are attached to two rigid blocks. One block is mounted on the surface and the other can move to get desired position and orientation. A Zed camera is hung over the setup to capture the configuration of the rod. The setup for 3D bending is the same as the setup for the block, which will be described in the next subsection.

\subsection{Modeling with boundary conditions: block}
\begin{figure}[ht]
\centering
\includegraphics[scale=0.15]{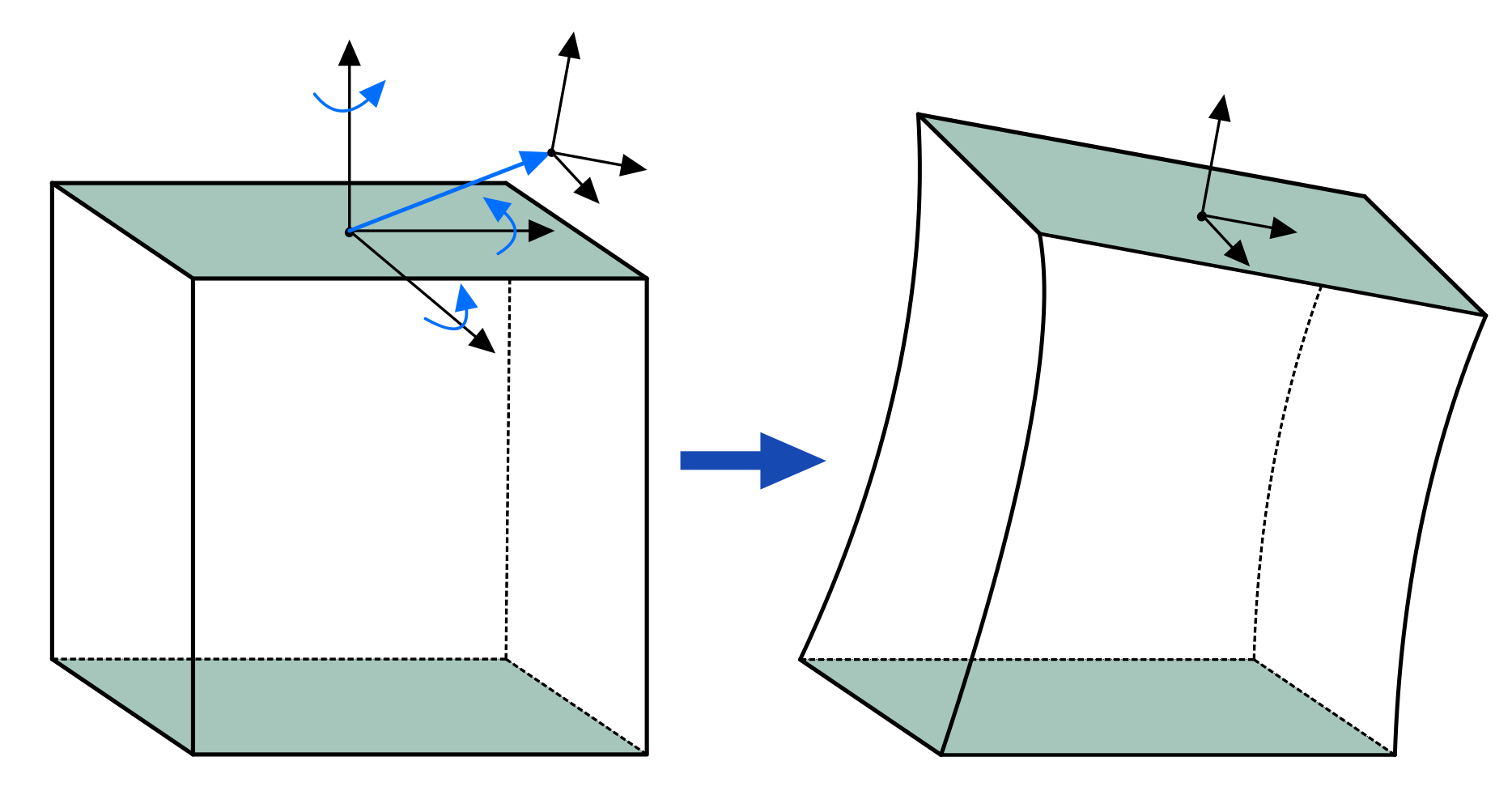}
\caption{Rubber block before and after deformation}
\label{defblock}
\end{figure}

In this section, a $3 cm \times 3 cm \times 6 cm$ rubber block is modeled to demonstrate the performance of the model with complex boundary conditions. 
Top and bottom surfaces are attached to rigid planes so that their shapes are preserved as the body deforms. A series of different top plane locations and orientations are then selected as the boundary conditions in this case study. 6 DoFs are identified as the top plane moves to desired positions and orientations, while the bottom plane remains fixed in the process, as shown in Figure \ref{defblock}.

The composition of locally volume-preserving primitives are used to model the deformation. The twist deformation gives the top plane freedom to rotate. Bending deformation involves rotation and displacement of the top plane. These two deformations provide the base case of modeling. Stretch deformation can change the arclength of the backbone curve. It is also essential to include shear deformation when constraints are on both position and orientation, as it allows the horizontal planes to slide along each other. Composition operation combines various deformations and generates a complex deformation with sufficient degrees of freedom while satisfying the boundary condition:
\begin{equation}
{\bf f(x)} = {\bf b}_2 {\bf (s(e(t(x))))}
\end{equation}
Note that bending primitive has to be composed last since it breaks the parallelism between horizontal planes.

Appropriate scaler functions are then selected in each primitive. Every individual primitive must satisfy the boundary condition that top and bottom surfaces are rigid. The scaler functions defining each deformation type are expanded as a weighted sum of modes in Table \ref{blockmode}.

\begin{table}[ht]
\caption{Modal expansions for modeling the rubber block}
\begin{center}
\begin{tabular}{c c c}
\hline
Deformation & Modal expansion \\
\hline
Twist and rotate & $ro+tw x_3$ \\
Stretch & $(x_3 (x_3-h) st+1)$\\
Shear($x_1$) & $s_1 x_3$ \\
Shear($x_2$) & $s_2 x_3$ \\
Bend & $ b_1 \sin(\frac{\pi x_3}{L})+b_2 \sin(\frac{2 \pi x_3}{L})+b_3 \sin(\frac{3 \pi x_3}{L})+b_4 \sin(\frac{4 \pi x_3}{L}) $ \\
\hline
\end{tabular}
\end{center}
\label{blockmode}
\end{table}

$ro, tw, st, s_1, s_2, b_1, b_2, b_3, b_4$ are the mode parameters. When using pure kinematics to model the deformation, it is important to reduce the number of redundant degrees of freedom. Only 9 DOFs are involved in the composite deformation, which keeps the modeling simple while representing sufficient DoFs.

\begin{figure}[ht]
\centering
\includegraphics[scale=0.05]{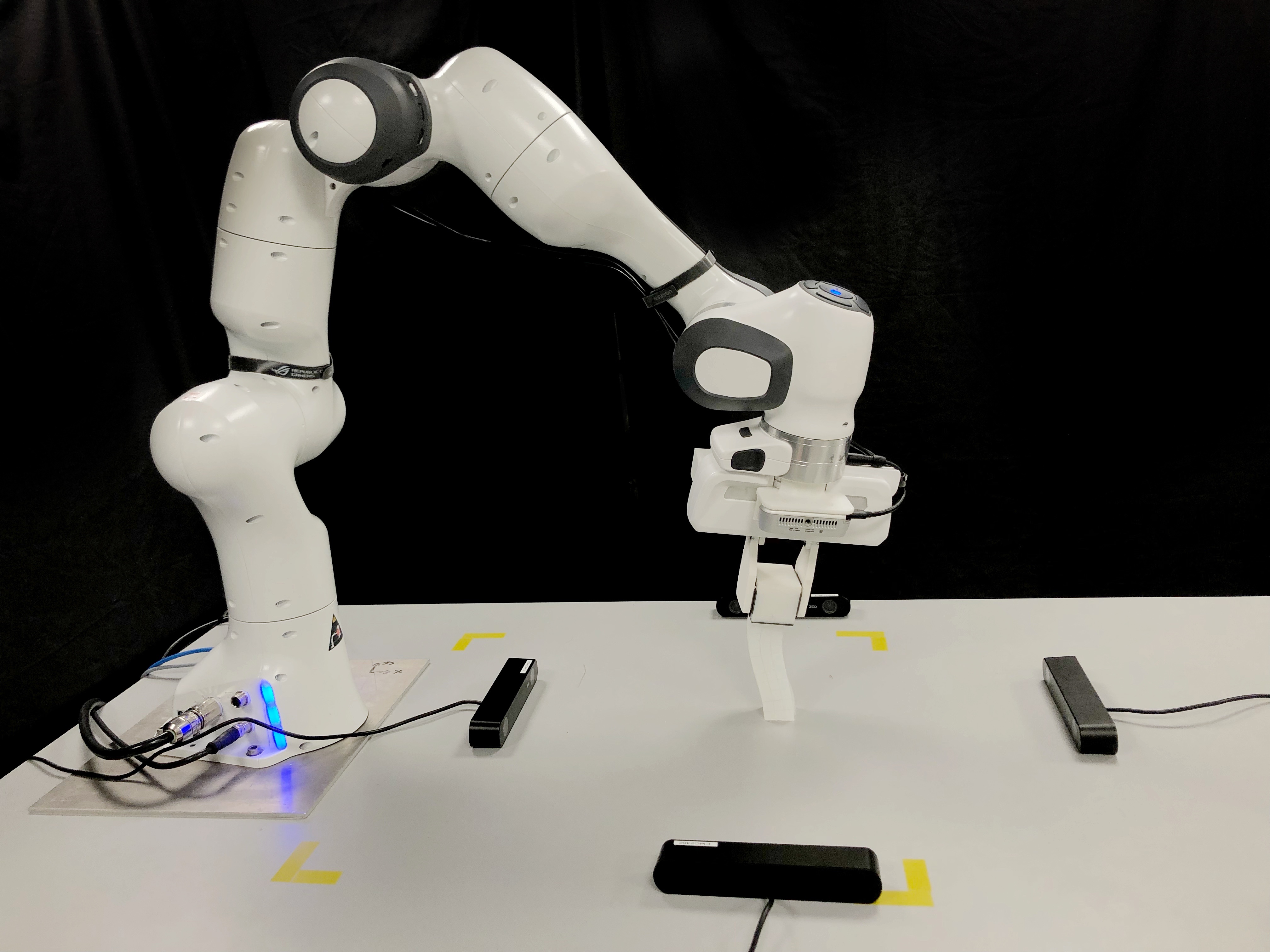}
\caption{Experiment setup of the block case: rigid-link robot is used to generate positional and orientational boundary conditions.}
\label{blocksetup}
\end{figure}
The experiment setup is shown in Figure \ref{blocksetup}. A rubber block is mounted on the table. A rigid handle is attached to the top surface of the block. It is grasped by a Franka robot arm, which enables precise movement of the surface. Four ZED 2 stereo cameras are located on the side of the block.

\section{Result and discussion}
This section presents the performance of our method by comparing with FEM and physical experiments.



The prerequisite of modeling the deformation using kinematics is that the simulation result is similar regardless of the constitutive models. To verify this condition, we conduct a series of FEM simulations in ABAQUS where the Mooney-Rivlin coefficients are varied. The coefficients used are listed in Table \ref{testsensitivity}. Detailed boundary conditions are listed in Table \ref{sensiboundary}. Note that the boundary conditions in these simulations are kinematics boundary conditions such as the displacement of the top plane, inflation volume of the chamber, as stated in the former section. The deformation will not be the same if mechanics boundary conditions are applied. For example: the inflation pressure of the chamber.
\begin{table}[ht]
\caption{The Mooney-Rivlin coefficients used to test the sensitivity of FEM to material property.}
\begin{center}
\begin{tabular}{c c c}
\hline
Material & $C_{10} (kPa)$ & $C_{01} (kPa)$ \\
\hline
Ecoflex 00-10 & 0.624 & 0.746 \\
Ecoflex 00-30 & 5.6 & 6.3 \\
Ecoflex 00-50 & 10.4 & 21.4 \\
Dragonskin 00-30 & 1.19 & 23.028 \\
\hline
\end{tabular}
\end{center}
\label{testsensitivity}
\end{table}

\begin{table}[ht]
\caption{The boundary conditions of the chamber, rod, and block case used to test the sensitivity of FEM to material property.}
\begin{center}
\begin{tabular}{c c c}
\hline
Case & Boundary condition type & Value \\
\hline
Chamber & Inflation volume & $105 mL$ \\
Rod & \makecell[c]{Displacement of the top plane along the $x,y,z$ axis; \\Rotation of the top plane about the $x,y,z$ axis.} & \makecell[c]{$1.5 cm, 1.5 cm, -3 cm$; \\$0.1 rad, 0.1 rad, 0.1 rad$.} \\
Block & \makecell[c]{Displacement of the top plane along the $x,y,z$ axis; \\Rotation of the top plane about the $x,y,z$ axis.} & \makecell[c]{$0.6 cm, 0.6 cm, 0.6 cm$; \\$0.1 rad, 0.1 rad, 0.1 rad$.}\\
\hline
\end{tabular}
\end{center}
\label{sensiboundary}
\end{table}

\begin{figure}[ht]
\centering
\includegraphics[scale=0.5]{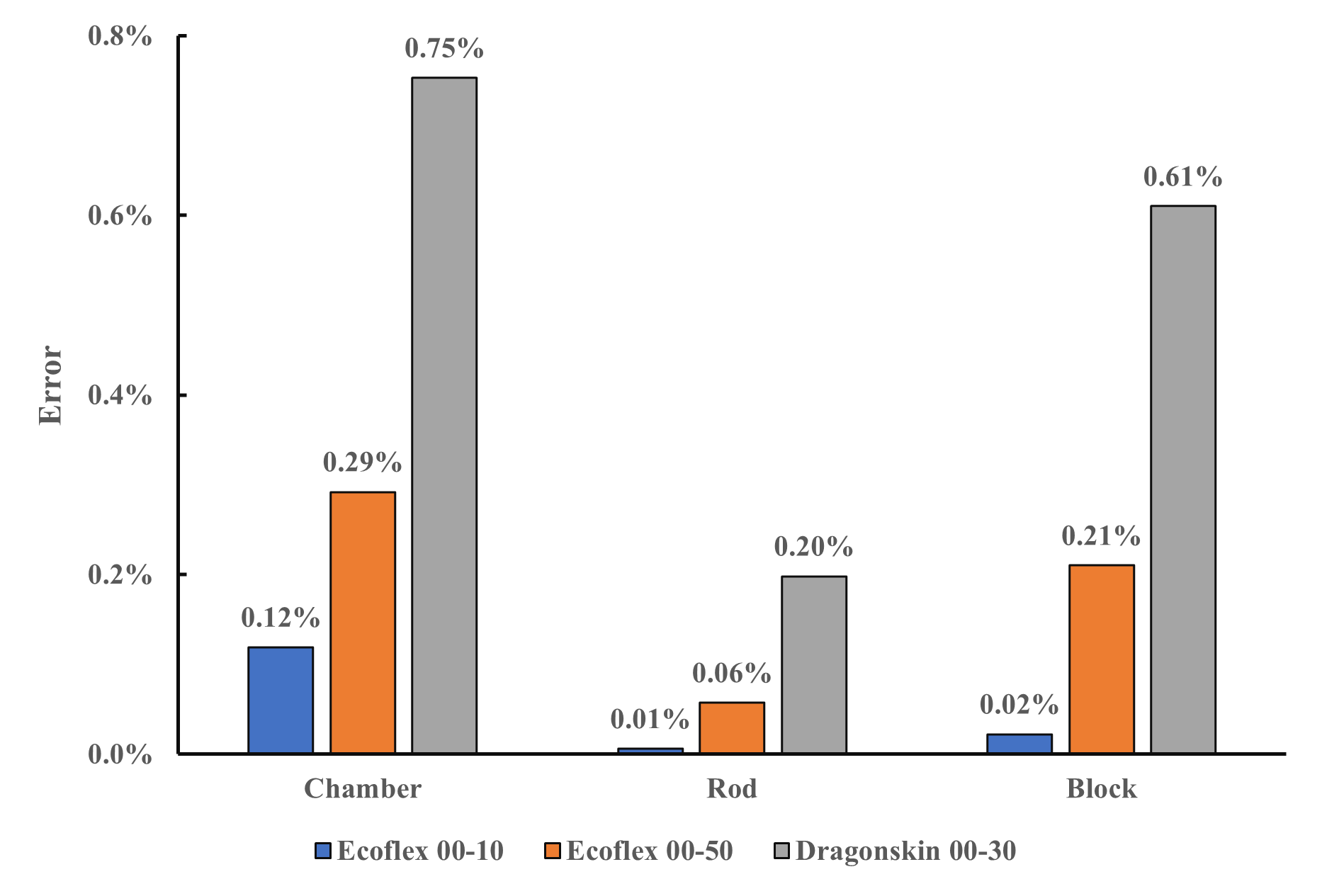}
\caption{The difference of simulation results between three different materials and Ecoflex 00-30.}
\label{sensitive}
\end{figure}

The simulation results of different materials are similar to each other. Use the result of Ecoflex 00-30 as the baseline, compare other results with it using (\ref{abacom}), the differences for all the materials are lower than $1\%$, as shown in Figure \ref{sensitive} The result proves that this condition holds for all the cases studied in this paper.

\begin{figure}
\centering
\includegraphics[scale=0.4]{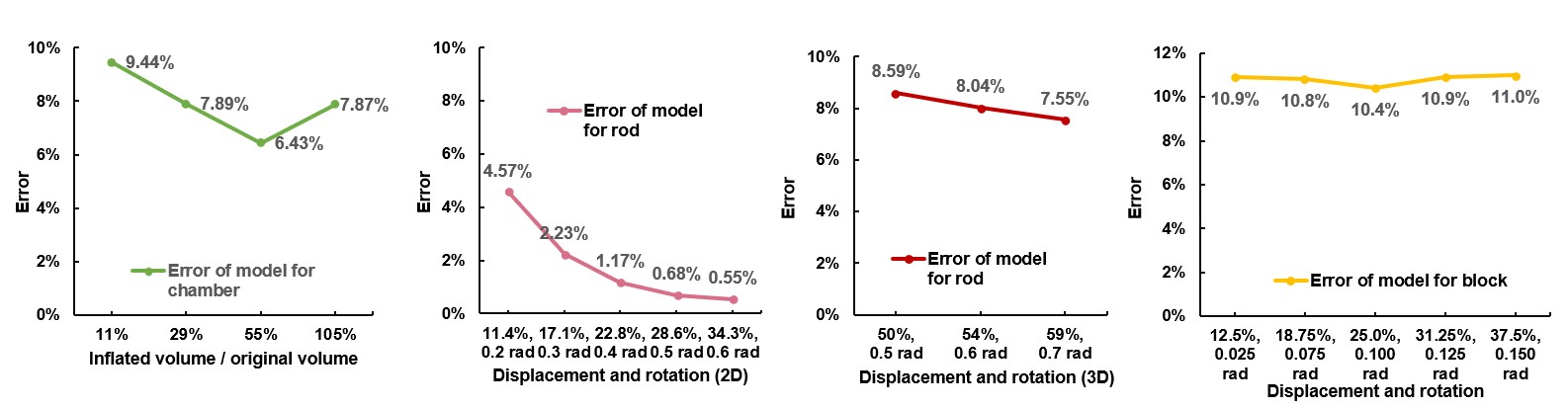}
\caption{The error of the model compare to ABAQUS when the degree of deformation varies. The percentage represents the ratio of displacement of the center point of the top plane to the dimension of the block, or the ratio of the volume change of the chamber to the original volume of the chamber.}
\label{compareabaqus}
\end{figure}

Figure \ref{compareabaqus} shows the percentage error between the primitives model and ABAQUS for different states and use cases. The primitives method stably presents minor deviation from ABAQUS even with a large displacement of the top plane. The percentage errors are controlled within 11$\%$ range in all cases. The rod bending in the 2D space case has a minimal error at $<5\%$, which might be resulted from the simplicity of the system. As a result, the modal expansion can fully specify the rod-like system. The percentage error also decreases with the increased displacement. This is attributed to the fact that the coordinates follow closely to the FEM as the rod becomes more distorted. The 3D bending case and chamber case also exhibit a similar descending trend as the deformation level increases. However, in the chamber case, this no longer holds with high inflation volume. This might be because our model does not consider the elongation and contraction in the $z$ direction, which cannot be neglected anymore when the system volume is overly large. The percentage error in the block case is the largest. This might be due to the bending primitive involved in this case is planar bending, which could be a significant simplification for the modeling of 3D deformation. Nonetheless, the error still falls within an acceptable range, which remains competitive with FEM considering the speed advantage of this method.

\begin{figure}[]
\centering
\includegraphics[width=0.9\textwidth]{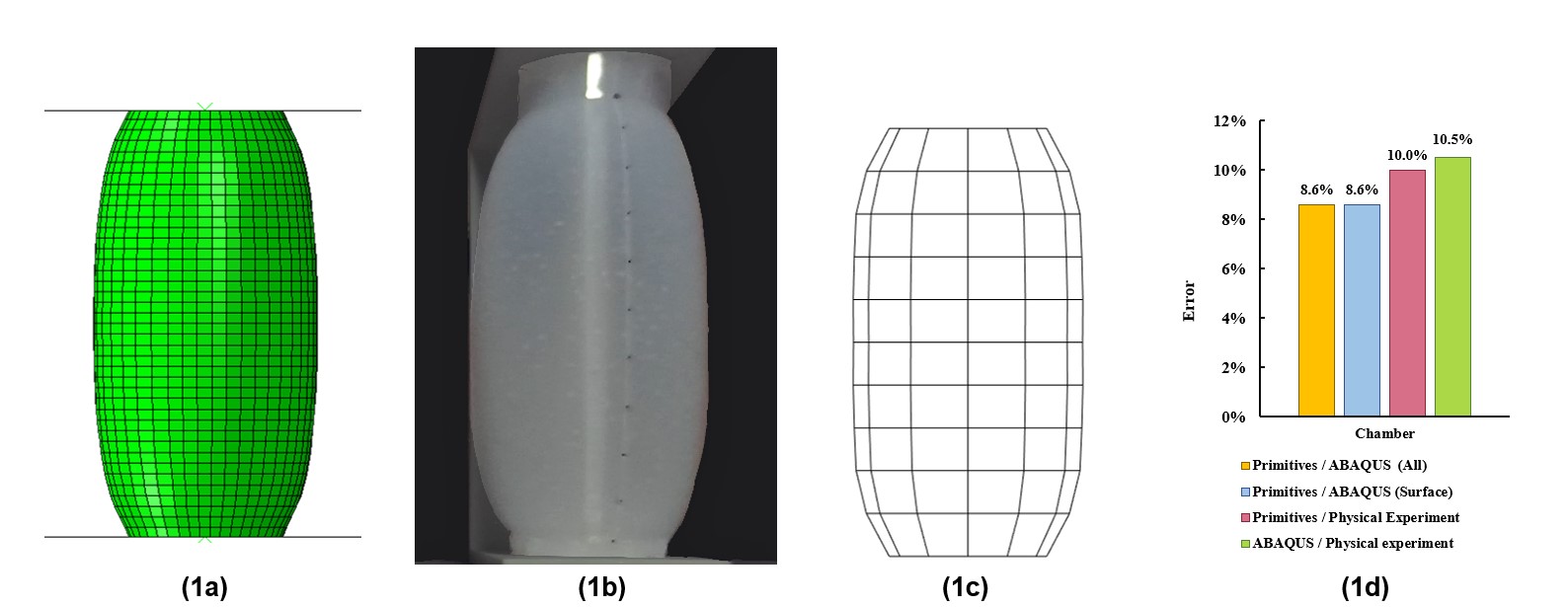}
\includegraphics[width=0.9\textwidth]{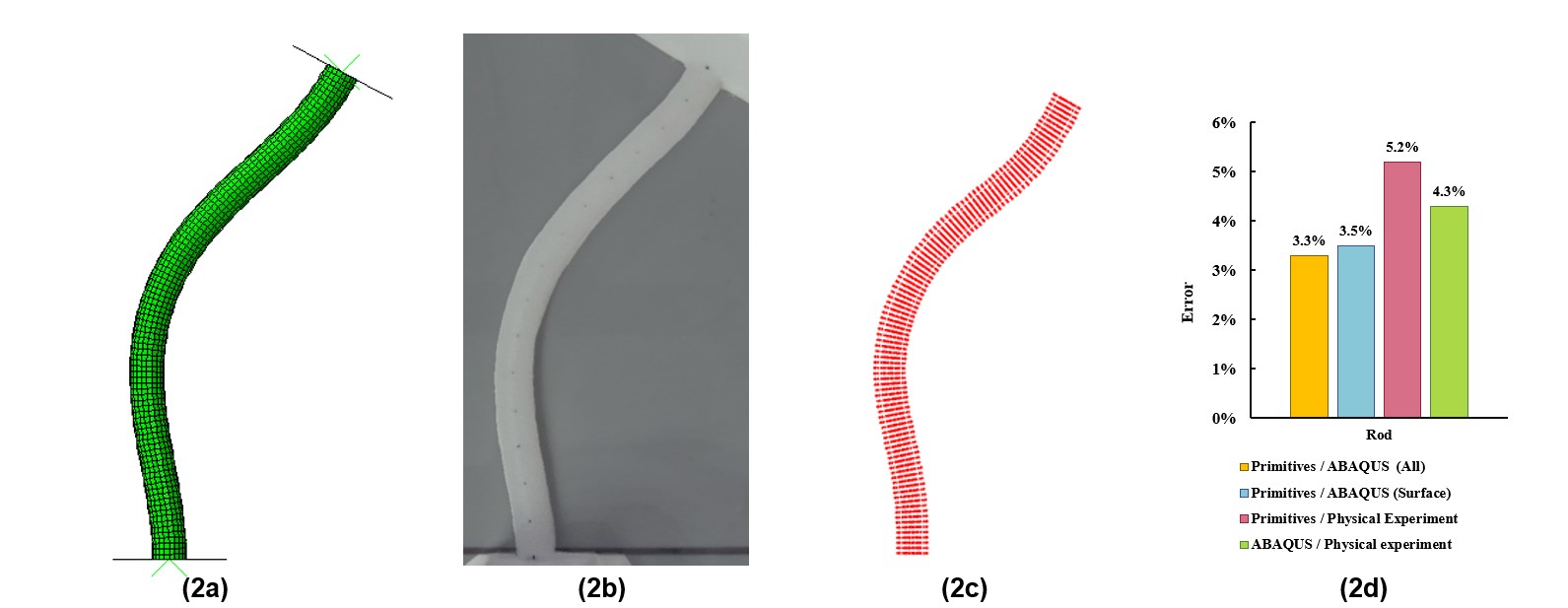}
\includegraphics[width=0.9\textwidth]{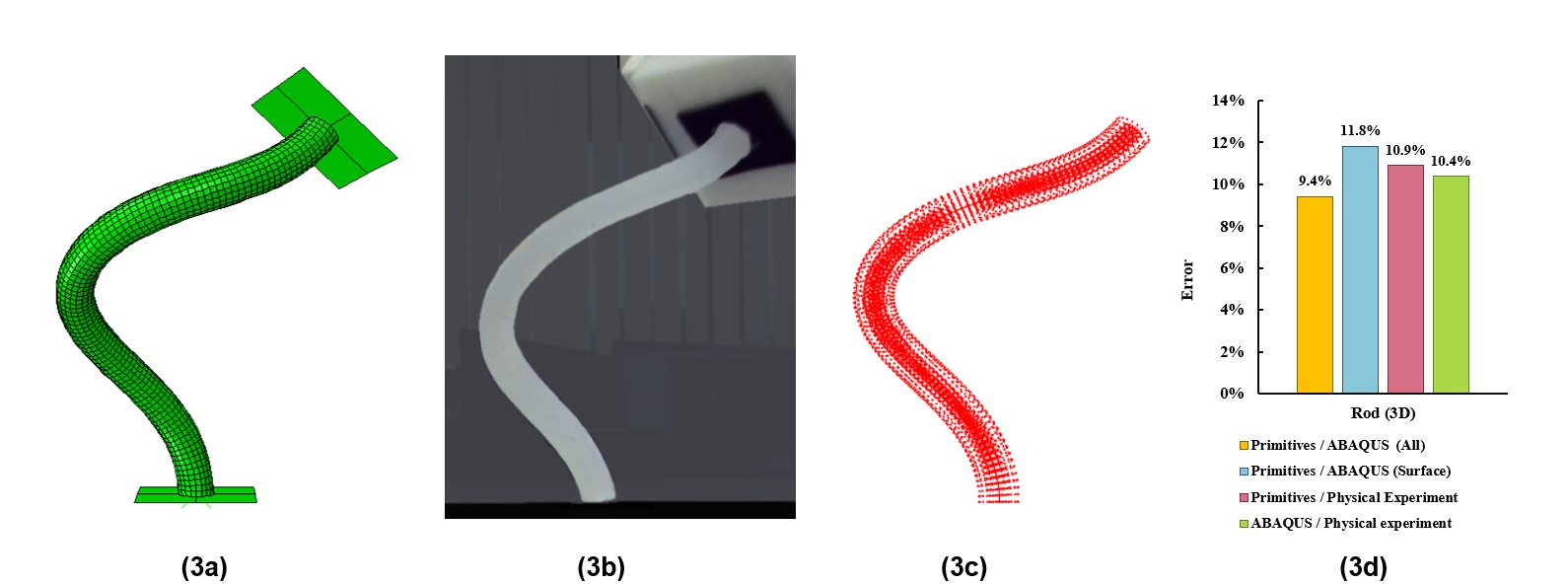}
\includegraphics[width=0.9\textwidth]{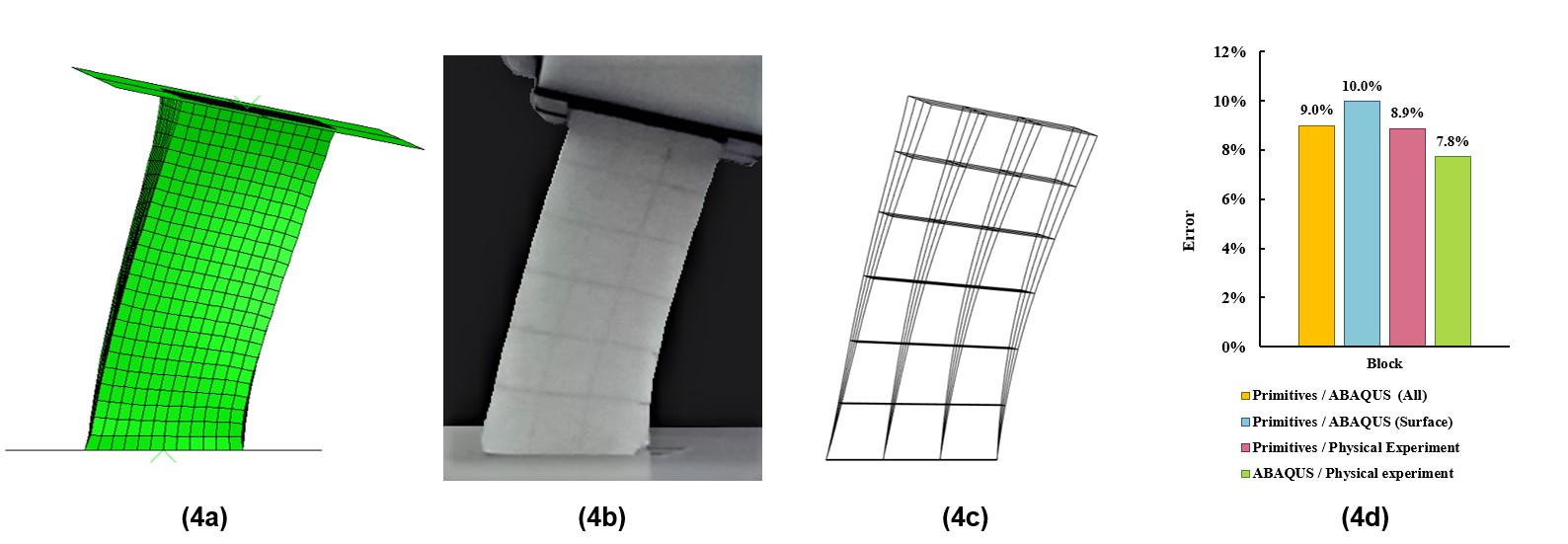}
\caption{The experiment result of 1. chamber; 2. rod in 2D; 3. rod in 3D; 4. block. (a) The simulation result from ABAQUS; (b) A photo shot by Zed camera during the experiment; (c) The simulation result from our method; (d) the comparison between two simulation results and experiment. The compare with physical experiment is at the surface. "All" in the legend means the compare between primitive model and ABAQUS model is about all the nodes.}
\label{threeresults}
\end{figure}

The primitives model and ABAQUS are also validated by physical experiments. Figure \ref{threeresults} shows the three-way comparison of the results in percentage error. Note that only the points on the object surface are involved in the comparison with the physical experiment, as the camera can only capture the coordinates of surface points. It can be observed that ABAQUS and primitives model have similar percentage errors when benchmarked against physical experiments. During the comparison with ABAQUS, we observe that our method has a larger error at the surfaces of the object. This is due to the assumption in our method that horizontal planer sections remain plane while the real objects usually not preserve these planes. Even so, our method still presents similar error compared to error from ABAQUS. This shows the scaler functions selected can accurately describe the deformation of these systems, which proves the capability of this method with proper modes selection.

\begin{figure}[]
\centering
\includegraphics[width=0.2\textwidth]{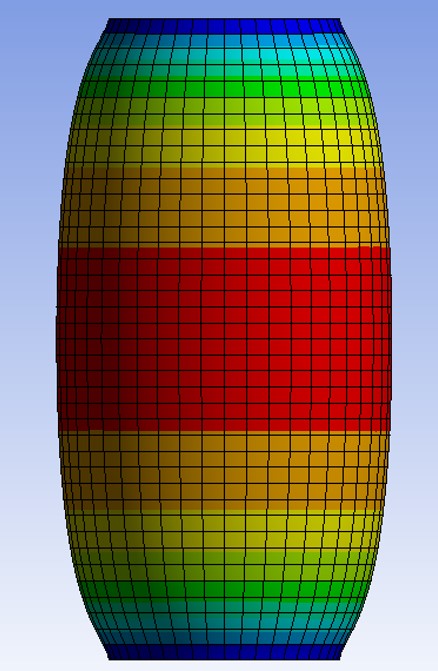}
\includegraphics[width=0.3\textwidth]{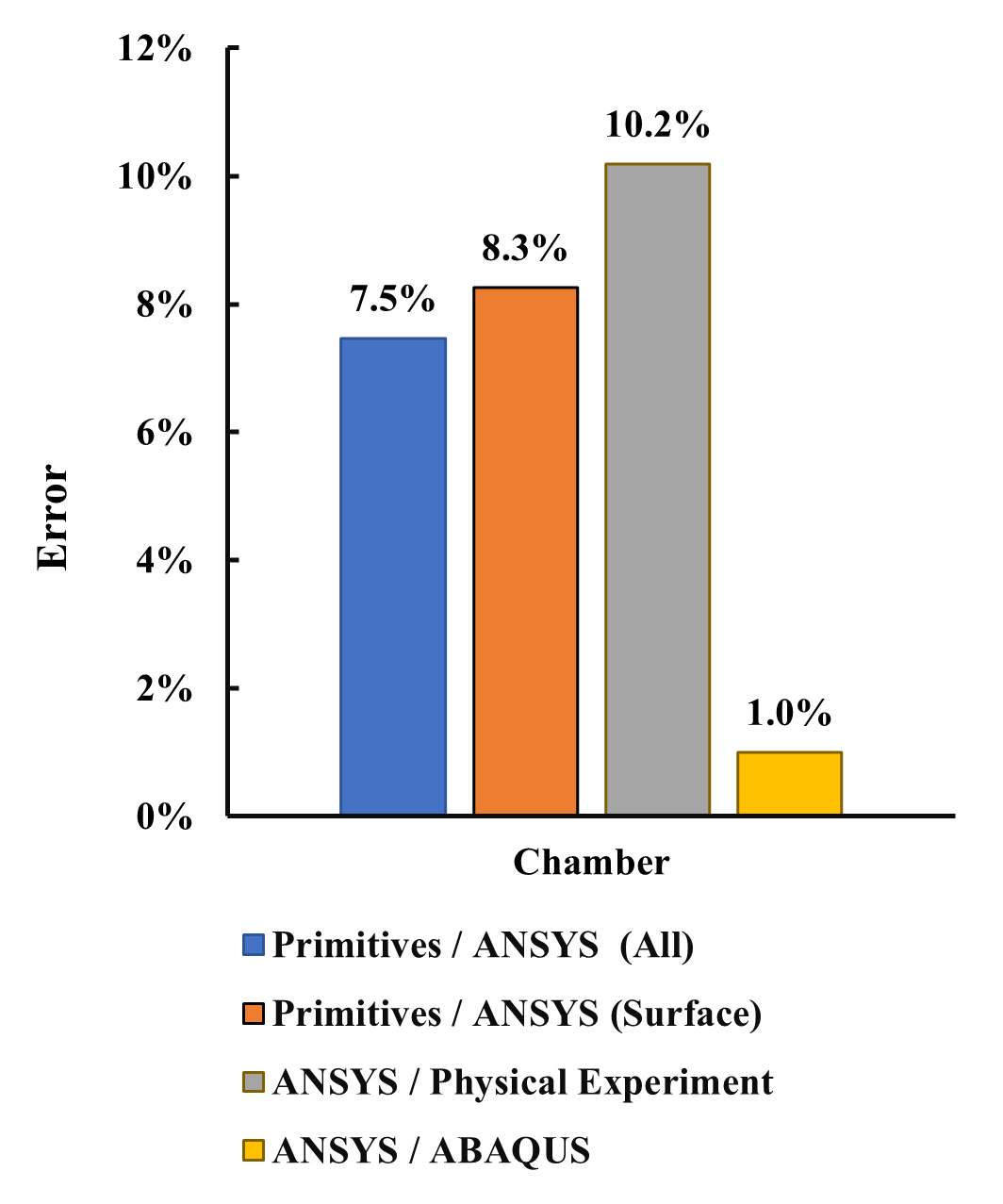}
\\
\includegraphics[width=0.23\textwidth]{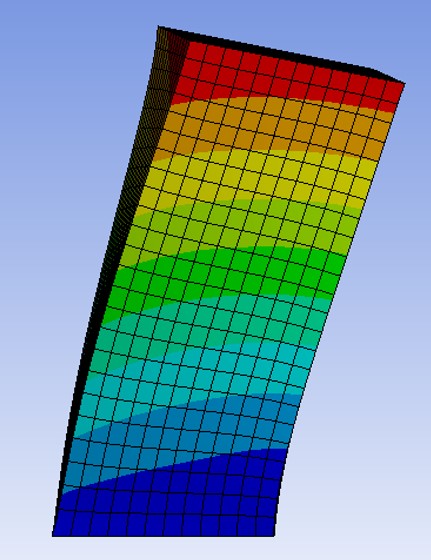}
\includegraphics[width=0.3\textwidth]{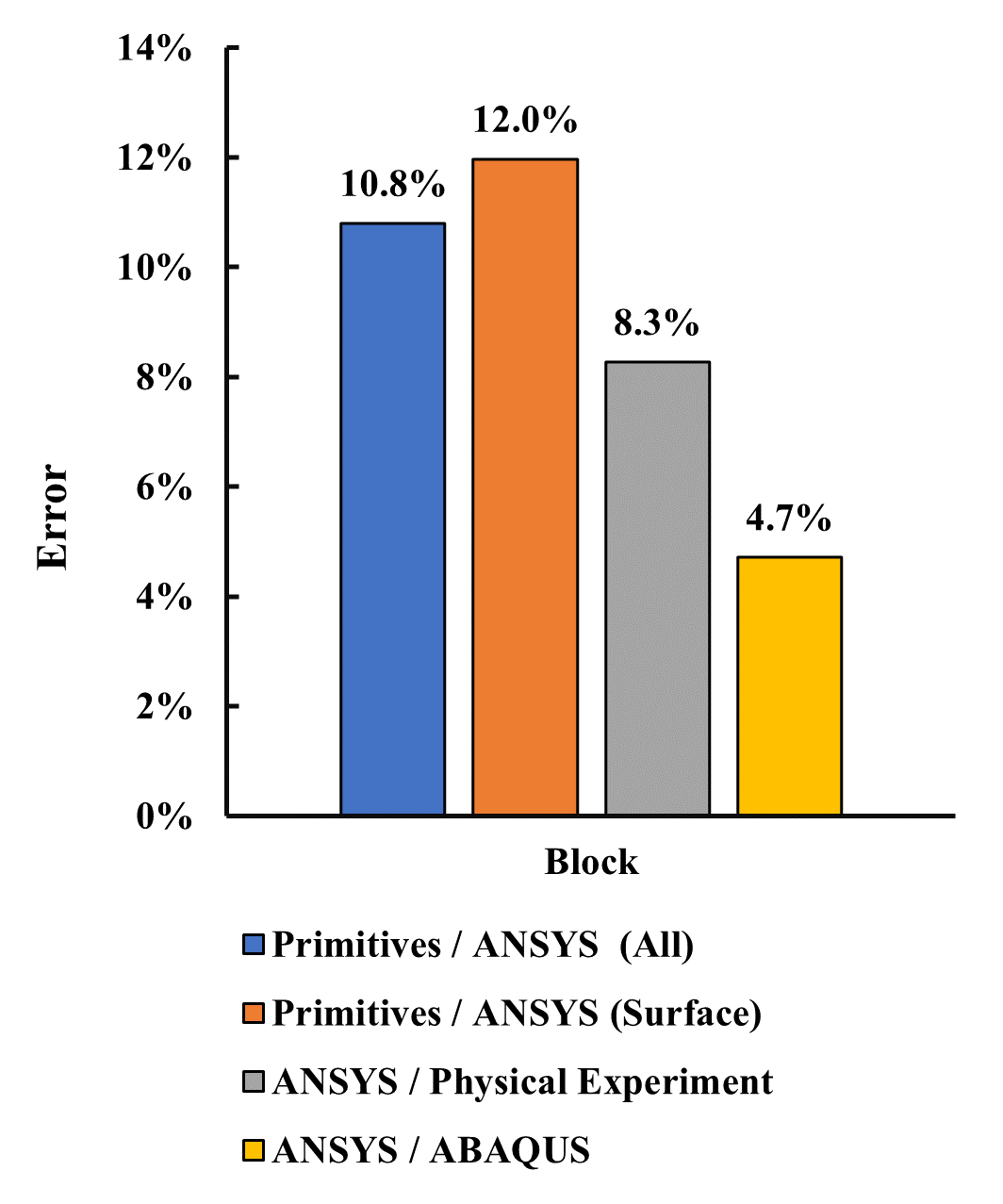}
\caption{The simulation result from ANSYS.}
\label{ansys}
\end{figure}

The modeling process of different FEM softwares can be different, hence the result might be different. Therefore, we also tested the chamber and block case using another software ANSYS to verify our modeling method. The rod case in ANSYS does not converge to a reasonal configuration, therefore not included in the discussion here. Figure \ref{ansys} shows the error of ANSYS result compared to ABAQUS, our primitive model and physical experiment. Comparing with Figure \ref{threeresults}, we could discover that ANSYS and ABAQUS results are close to each other. But when compared to physical experiment, all three methods provide results with similar error.

\begin{figure}
\centering
\includegraphics[scale=0.7]{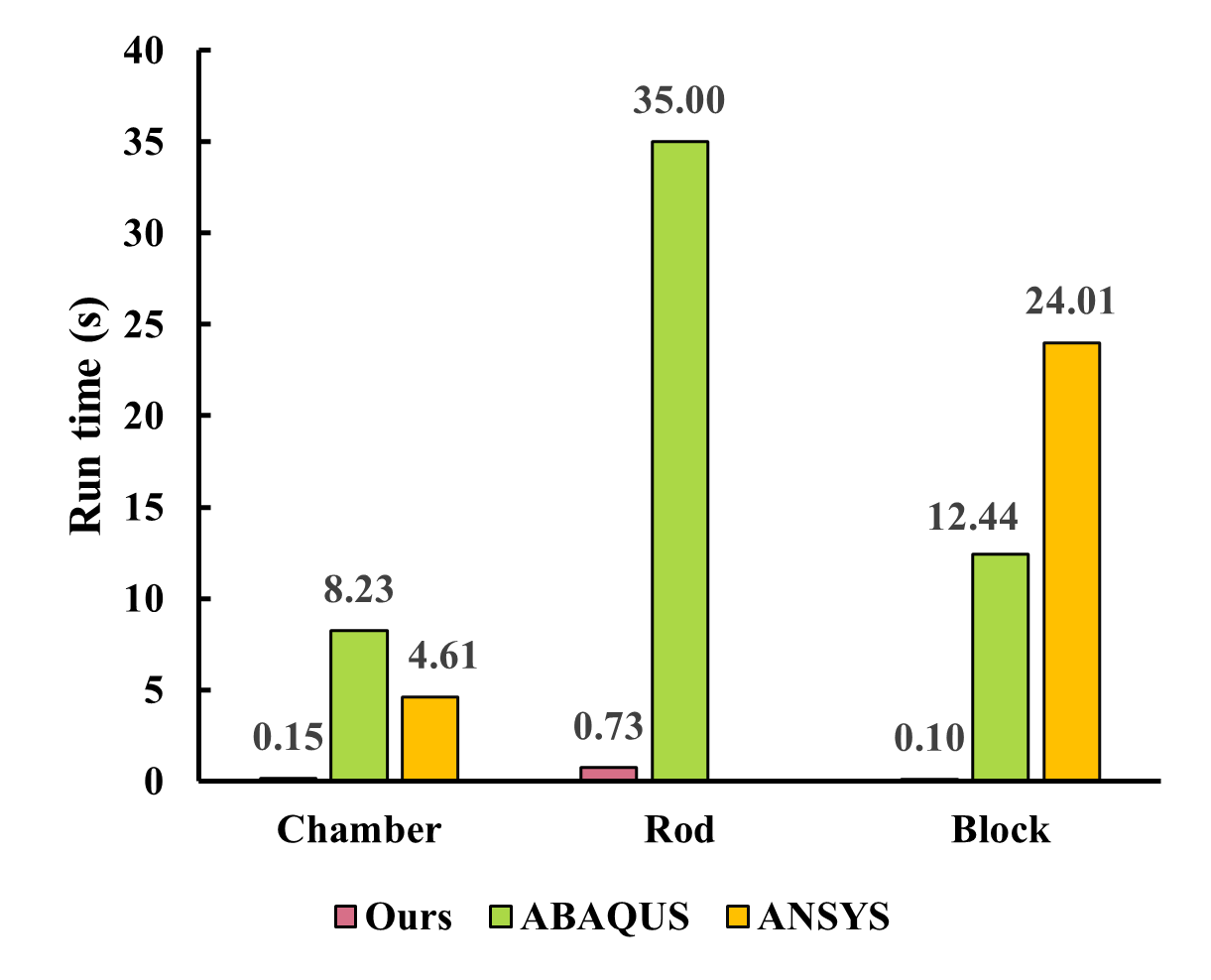}
\caption{The run time of ABAQUS, ANSYS and our method for the chamber, rod, and block case. The ABAQUS and ANSYS run time is the total CPU time, and the time of our method is the time required to compute all the mode parameters.}
\label{speed}
\end{figure}

The above results show that our method renders comparable coordinates as FEM. Nonetheless, our method is significantly faster. As illustrated in Figure \ref{speed}, the primitive method can be 100 times faster when simulating the deformation of a chamber. Similar speed differences are also observed in other cases. FEM needs to formulate global matrices of stiffness and forces and solve them, which generally takes the time complexity of $O(NW^2)$, where $N$ is the number of nodes, and $W$ is the bandwidth of the stiffness matrix \cite{farmaga2011evaluation}. In most cases, at least thousands of nodes are required to form a mesh with sufficient object representation, which results in the extended running time of FEM. In contrast, the primitive method does not involve complex mechanics computation during the optimization and it also does not need nodes. Instead, a closed-form solution is generated to describe all the points within the deformation field. The main factors that affect the speed of the primitive method are the number of modes, the number of iterations and the number of boundary conditions. In practice, less than 10 modes and 100 iterations would be sufficient for the modeling. No nodes selection is involved in this method, which means that the computing time scales well with the input size.

\begin{figure}
\centering
\includegraphics[width=1\textwidth]{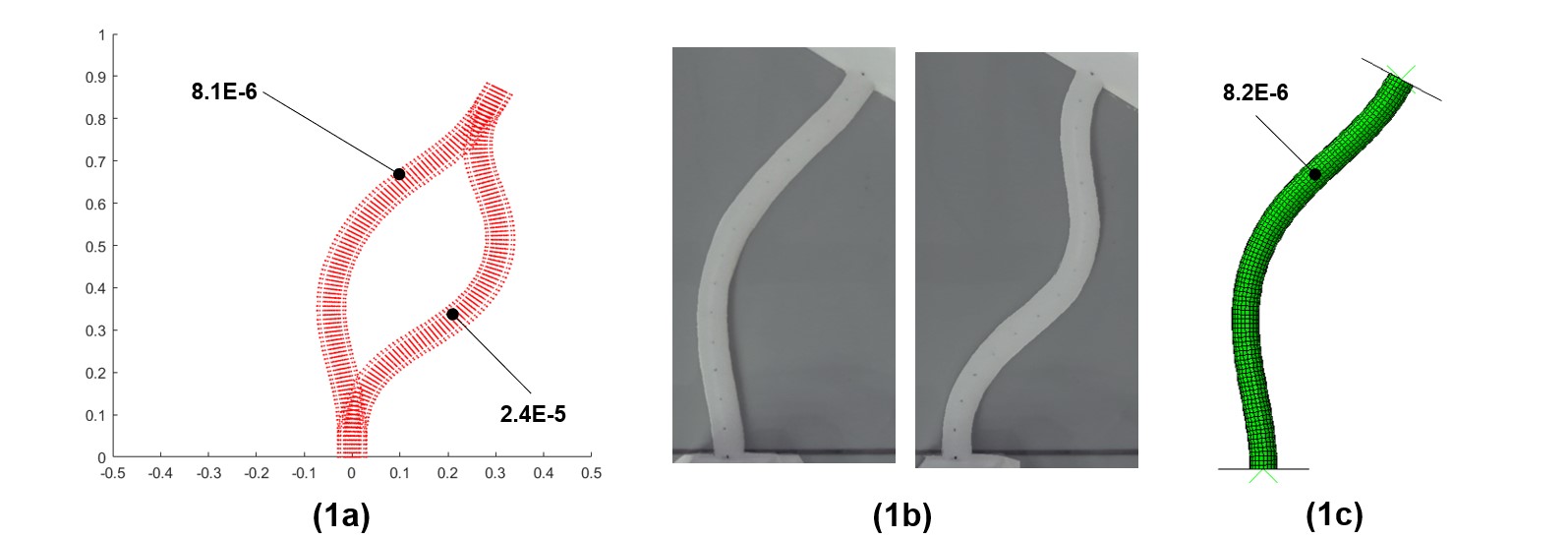}
\includegraphics[width=1\textwidth]{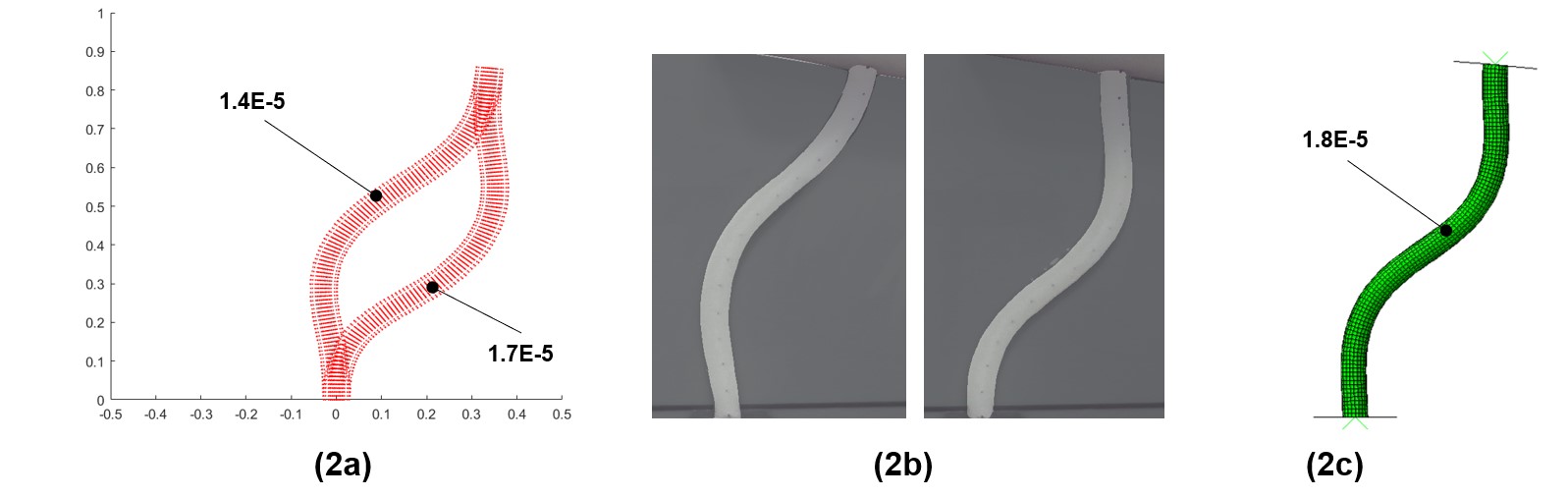}
\includegraphics[width=1\textwidth]{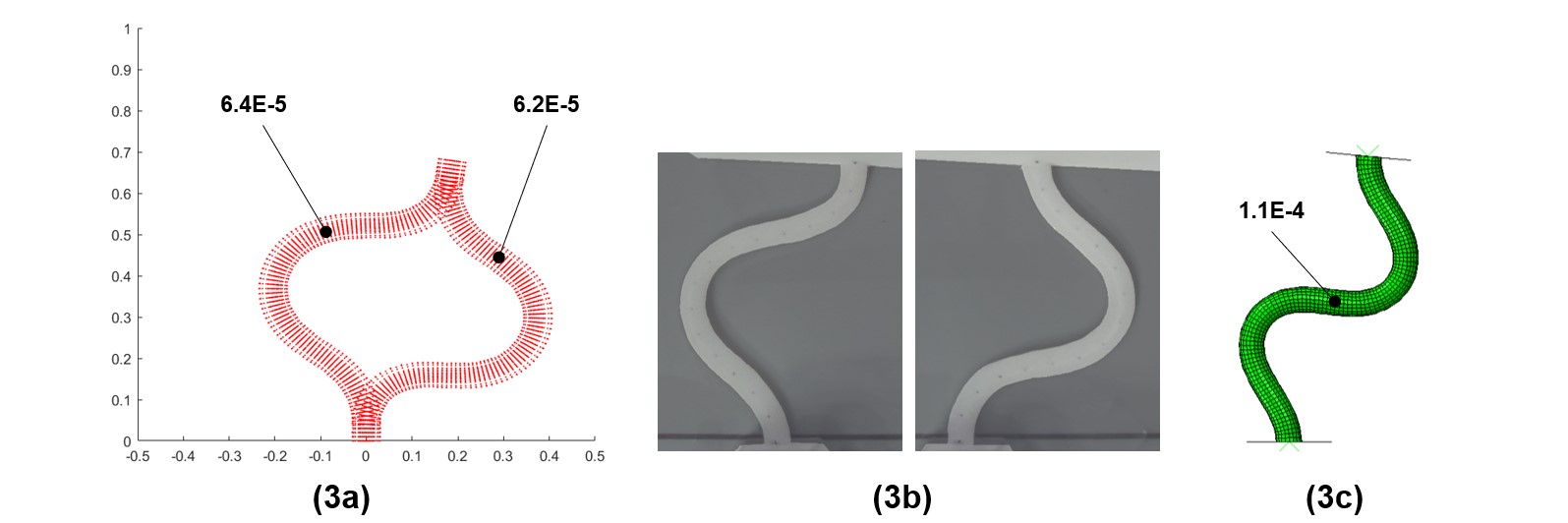}
\caption{The configuration generated by ABAQUS and our method. $a$ is the result from primitive model, $b$ is the upper-hand and lower-hand configuration of a real rod in the physical experiment, and $c$ is the result from ABAQUS. The strain energy of these simulation results are annotated in the Figure. Series 1 shows the case when ABAQUS and our method converge to the same branch. Our method also generates results from another branch. Series 2 presents the case when ABAQUS gives the higher energy branch result. Series 3 shows the case when ABAQUS is trapped in local minimum with significant error, while our approach still generates both branches with low error.}
\label{comparerods}
\end{figure}
Another phenomenon is observed during the rod case. The inverse kinematics problem for a redundant manipulator usually consists of a series of solution branches. Our modeling method could generate results from different solution branches, while ABAQUS only gives one result. Besides, primitive model has results with lower strain energy compared to ABAQUS, showing the optimization in our method performs better in finding a local minimum. Figure \ref{comparerods} gives a few examples when our method and ABAQUS give the same or different solution branches.

The results show that our method could be a time-efficient alternative to FEM. It performs similarly to FEM in terms of coordinates generation but it is about 50 times faster on average. To further improve the accuracy of our method, we should take into account the impact of gravity. Automated tools can also be developed to select the mode functions more efficiently by setting up the standard of selection. This will improve the reproducibility and reliability of this method. Post-processing techniques can be invented to finetune the model for more realistic results. For instance, a new layer of small deformation could be composed on top of the current deformation. The new layer could be tasked to minimize the strain energy of the system, which is not explicitly considered in the current method.

\section{Conclusion}
This paper introduces a modeling method using closed-form locally volume-preserving deformation primitives to model the deformation of soft structures when boundary conditions are applied, as is often the case in soft robotics problems. This method simulates the robot deformation at a global level, and it generates the coordinates in terms of closed-form deformations. The modeling technique is tested on three use cases: an inflatable chamber, a rod, and a block. The rod and chamber cases show that this method has potential application in the modeling of snake-like soft robots and pneumatic soft robots. The block case proves that our technique is capable of handling boundary conditions with high DoFs (6 in this case). In all these cases, our method is at least 50 times faster than the finite element method, while providing comparable results with an error of $10\%$ compared to FEM and physical experiment. This technique could thus be a reduced-order tool to store and reconstruct the output data from FEM, or a time-efficient alternative to FEM in real-time control.

\section*{Acknowledgment}
This work was supported by NUS Startup grants R-265-000-665-133, R-265-000-665-731,
Faculty Board account C-265-000-071-001, MOE grant R-265-000-655-114, and National Research Foundation, Singapore, under its Medium Sized Centre Programme - Centre for Advanced Robotics Technology Innovation (CARTIN), subaward R-261-521-002-592.

\section*{Appendix A}
This section derives the 3D bending primitive. Given a 3D bending deformation in this form:
\begin{equation}
{\bf b}_3({\bf x}) = {\bf a}(x_3) + \nu(x_1,x_2,x_3) {\bf n}(x_3) + \beta(x_1,x_2,x_3) {\bf b}(x_3) \,.
\label{bendtwist}
\end{equation}
In other words, the $x_3$ axis is deformed to become a 3D space curve parameterized by $x_3$ (which becomes arclength), and a transformed version of the $x_2-x_1$ plane is moved (and deformed) to reside in the ${\bf n}(x_3)-{\bf b}(x_3)$ plane. We specify ${\bf a}(x_3)$ and ask when it is possible for
$\nu(x_3,x_2,x_1)$ and $\beta(x_3,x_2,x_1)$ to be defined such that the volume is locally preserving. The first column of this Jacobian matrix is
written as
\begin{eqnarray*}
\frac{\partial {\bf b}_3}{\partial x_3} \,&=&\, {\bf a}'(x_3) + \frac{\partial \nu}{\partial x_3}{\bf n}(x_3)
+ \nu(x_1,x_2,x_3){\bf n}'(x_3) + \frac{\partial \beta}{\partial x_3}{\bf b}(x_3)
+ \beta(x_1,x_2,x_3){\bf b}'(x_3)\\
\,&=&\, {\bf t}(x_3) + \frac{\partial \nu}{\partial x_3}{\bf n}(x_3) + \nu(x_1,x_2,x_3)(-\kappa(x_3) {\bf t}(x_3) + \tau(x_3) {\bf b}(x_3))\\
&\quad& + \frac{\partial \beta}{\partial x_3}{\bf b}(x_3) + \beta(x_1,x_2,x_3)(-\tau(x_3) {\bf n}(x_3)) \\
\,&=&\, (1-\kappa \nu){\bf t} + (\frac{\partial \nu}{\partial x_3} - \tau \beta){\bf n}
+ (\frac{\partial \beta}{\partial x_3} + \tau \nu){\bf b} \,.
\end{eqnarray*}

The second and third columns are
$$ \frac{\partial {\bf b}_3}{\partial x_2} \,=\,
\frac{\partial \nu}{\partial x_2} {\bf n} + \frac{\partial \beta}{\partial x_2} {\bf b} $$
and
$$ \frac{\partial {\bf b}_3}{\partial x_1} \,=\,
\frac{\partial \nu}{\partial x_1} {\bf n} + \frac{\partial \beta}{\partial x_1} {\bf b} \,. $$

Then the Jacobian matrix can be written as

\begin{equation}
\nabla_{\bf x} {\bf b}_3\,=\, R \,
\left(\begin{array}{ccc}
(1-\kappa \nu) & 0 & 0 \\ \\
(\frac{\partial \nu}{\partial x_3} - \tau \beta) & \frac{\partial \nu}{\partial x_2} & \frac{\partial \nu}{\partial x_1} \\ \\
(\frac{\partial \beta}{\partial x_3} + \tau \nu) & \frac{\partial \beta}{\partial x_2} & \frac{\partial \beta}{\partial x_1}
\end{array}\right),
\label{ksssdfdee}
\end{equation}
where $R=[\bf t,n,b]$.
Since $\det(AB) = \det(A) \det(B)$ and since $\det R = +1$, the deformation ${\bf f}({\bf x})$ will be locally volume preserving
if and only if
\begin{equation}
(1-\kappa \nu)
\left|\begin{array}{ccc}
\frac{\partial \nu}{\partial x_2} & \frac{\partial \nu}{\partial x_1} \\ \\
\frac{\partial \beta}{\partial x_2} & \frac{\partial \beta}{\partial x_1} \end{array}\right| \,=\,1.
\label{finalcondition}
\end{equation}

Suppose that $\beta$ is only a function of $x_3$,$x_1$, then
\begin{equation}
(1-\kappa(x_3) \nu(x_1,x_2,x_3)) \frac{\partial \beta(x_1,x_3)}{\partial x_1} \frac{\partial \nu(x_1,x_2,x_3)}{\partial x_2} \,=\, 1
\label{solvable1}
\end{equation}
Integrate this equation with respect to $x_2$:
\begin{equation}
\frac{\partial \beta}{\partial x_1} \nu - \frac{1}{2} \frac{\partial \beta}{\partial x_1} \kappa \nu^2 = x_2 + c(x_1,x_3)
\label{quadratic}
\end{equation}
$c(x_1,x_3)$ is an arbitrary function that corresponds to a shear deformation in normal direction of the backbone curve. Then solve for $\nu(x_1,x_2,x_3)$:
\begin{equation}
\nu(x_1,x_2,x_3)=\frac{\frac{\partial \beta}{\partial x_1} \pm \sqrt{\frac{\partial \beta}{\partial x_1}^2 - 2 \frac{\partial \beta}{\partial x_1} \kappa (x_2+ c(x_1,x_3))}}{\frac{\partial \beta}{\partial x_1} \kappa}
=\frac{1 \pm \sqrt{1 - \frac{2 \kappa (x_2+ c(x_1,x_3))}{\frac{\partial \beta}{\partial x_1}}}}{ \kappa}
\label{nu}
\end{equation}
When $\kappa(x_3)$ goes to zero, $\nu(x_1,x_2,x_3)$ should converge to $x_2$ to make sure the undeformed configuration matches with the original one. Hence, the negative root is chosen and when $\kappa(x_3)$ goes to zero, $\frac{\partial \beta}{\partial x_1}$ should goes to 1 and $c(x_1,x_3)$ should goes to zero. Besides, in order to avoid singularities, we should limit $\frac{\frac{\partial \beta}{\partial x_1}}{2 \kappa}-c > x_2$.

The simplest example is let:
\begin{equation}
\beta=x_1,
\label{simplebeta}
\end{equation}
Substitute (\ref{simplebeta}) into (\ref{nu}) and get:
\begin{equation}
\nu=\frac{1 - \sqrt{1 - 2 \kappa x_2}}{ \kappa}.
\label{simplenu}
\end{equation}
Then (\ref{bendtwist}) becomes:
\begin{equation}
{\bf b}_3({\bf x}) = {\bf a}(x_3) + \frac{1 - \sqrt{1 - 2 \kappa x_2}}{ \kappa} {\bf n}(x_3) + x_1 {\bf b}(x_3) \,.
\label{bendtwistnew}
\end{equation}
In this case, if the backbone curve is a planer curve, then the 3D bending becomes 2D bending.

There can be other choices. Suppose $\nu$ is only a function of $x_3$,$x_2$, then
\begin{equation}
(1-\kappa(x_3) \nu(x_2,x_3)) \frac{\partial \beta(x_1,x_2,x_3)}{\partial x_1} \frac{\partial \nu(x_2,x_3)}{\partial x_2} \,=\, 1
\label{solvable2}
\end{equation}
Integrate this equation with respect to $x_1$:
\begin{equation}
(1-\kappa \nu) \frac{\partial \nu}{\partial x_2} \beta \,=\, x_1 + c(x_2,x_3)
\label{linear}
\end{equation}
Solve for $\beta(x_1,x_2,x_3)$:
\begin{equation}
\beta (x_1,x_2,x_3)\,=\, \frac{x_1+ c(x_2,x_3)}{(1-\kappa \nu) \frac{\partial \nu}{\partial x_2}}
\label{linear}
\end{equation}
Similarly, when $\kappa(x_3)$ goes to zero, $\frac{\partial \nu}{\partial x_2}$ should goes to 1, $c(x_2,x_3)$ should goes to zero.

\section*{Appendix B}
\begin{figure}
\centering
\includegraphics[scale=0.35]{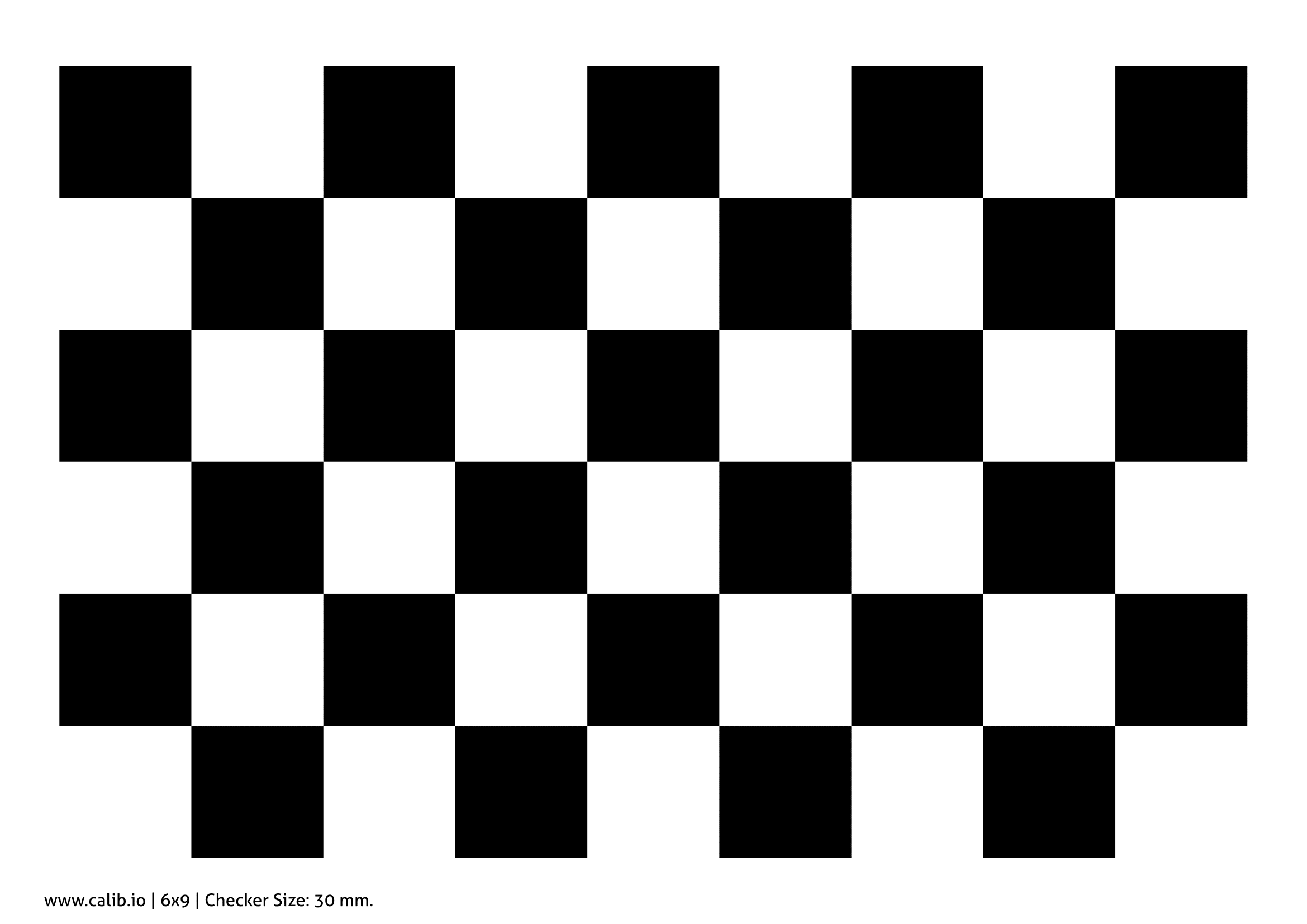}
\caption{This pattern is generated by www.calib.io. The width of the checker is 30 $mm$.}
\label{checkerboard}
\end{figure}

This section introduces the calibration process of the Zed cameras. Zed takes 2 images from different angles at the same time to allow the stereo matching. Zhang's method \cite{zhang2000flexible} is used to calibrate the camera. A checkerboard is used as the calibration gauge. It is printed on A4 paper and then sticked on a flat rigid surface. The checkerboard used to calibrate the camera is shown in Figure \ref{checkerboard}.

Zed takes 15 pairs of images of the board at different angles. These images are then fed into the stereo calibration function in Matlab computer vision toolbox. The calibration is done automatically and gives the intrinsic, extrinsic, and lens distortion parameters of the stereo camera. With these parameters, the undistorted image can be obtained for further measurement using the undistortedIimages function. The reprojection error of the calibration is below 0.5 pixel.

After obtaining the undistorted image, the 3D coordinate of any point that is included in both images can be reconstructed by triangulation. Matlab function triangulate could compute the 3D position of matching pairs of points on the image.

\begin{table}[ht]
\caption{The length measured by Zed camera.}
\begin{center}
\begin{tabular}{c c c}
\hline
Real length $(mm)$ & Zed measurement $(mm)$ & Average error \\
\hline
& 49.3142 & \\
50 & 49.5721 & $1.05\%$ \\
& 49.5435 & \\
& 113.9308 & \\
115 & 113.8508 & $0.93\%$ \\
& 114.0061 & \\
\hline
\end{tabular}
\end{center}
\label{testaccuracy}
\end{table}

We use a known object to test the accuracy of the measurement. A 3D-printed object of 50 $mm$ width and 115 $mm$ height is placed in front of the calibrated zed camera. The camera takes left and right images of the object and measures the height and width of it. Then compare the measurement result with the true dimensions of the object. The result is listed in Table \ref{testaccuracy}. The error is about 0.99$\%$.


\end{document}